\documentclass[journal]{IEEEtran}
\usepackage[cmex10]{amsmath}
\usepackage{times}
\usepackage{epsfig}
\usepackage{graphicx}
\usepackage{amssymb}
\usepackage{algorithmic}
\usepackage{algorithm}
\usepackage{color}
\usepackage{subfigure}
\usepackage{tikz}
\usepackage{mathrsfs}
\usepackage{makecell}
\usepackage{bm}
\usepackage{multicol}
\usepackage{multirow}
\usepackage[numbers,sort&compress]{natbib}
\usepackage[section]{placeins}
\usepackage{url}
\usepackage{array}
\definecolor{fgreen}{RGB}{34,139,34}

\def\eg{{e.g.}}
\def\ie{{i.e.}}
\def\etal{\emph{et al. }}
\def\etc{{etc.}}

\definecolor{myblue}{RGB}{20,50,200}
\definecolor{mygreen}{RGB}{34,139,34}

\begin{document}
\title{Multiple Instance Dictionary Learning for Beat-to-Beat Heart Rate Monitoring from Ballistocardiograms}

\author{ Changzhe Jiao,~\IEEEmembership{Student Member,~IEEE,} Bo-Yu Su,~\IEEEmembership{Student Member,~IEEE,} Princess Lyons,~\IEEEmembership{Student Member,~IEEE,} Alina Zare,~\IEEEmembership{Senior Member,~IEEE}, K. C. Ho,~\IEEEmembership{Fellow,~IEEE},  and Marjorie Skubic,~\IEEEmembership{Senior Member,~IEEE} \thanks{Manuscript received June 14, 2017; revised January 6, 2018 and February 15, 2018; accepted February 25, 2018. This material is based upon work supported by the National Science Foundation under Grant No. IIS-1350078 - CAREER: Supervised Learning for Incomplete and Uncertain Data.} \thanks{Changzhe Jiao is with the Key Laboratory of Intelligent Perception and Image Understanding, Ministry of Education of China, School of Artificial Intelligence, Xidian University, Xi'an 710071, China; Bo-Yu Su, K. C. Ho,  and Marjorie Skubic are with the Department of Electrical Engineering and Computer Science, University of Missouri, Columbia, MO, 65211 USA; Princess Lyons and Alina Zare are with the Department of Electrical and Computer Engineering, University of Florida, Gainesville, FL, 32611 USA (correspondence e-mail: azare@ufl.edu, cjiao@mail.missouri.edu).   }
}

\markboth{}{}

\maketitle

\begin{abstract}
	A multiple instance dictionary learning approach, Dictionary Learning using Functions of Multiple Instances (DL-FUMI), is used to perform beat-to-beat heart rate estimation and to characterize heartbeat signatures from ballistocardiogram (BCG) signals collected with a hydraulic bed sensor. DL-FUMI estimates a ``heartbeat concept'' that represents an individual's personal ballistocardiogram heartbeat pattern. DL-FUMI formulates heartbeat detection and heartbeat characterization as a multiple instance learning problem to address the uncertainty inherent in aligning BCG signals with ground truth during training. Experimental results show that the estimated heartbeat concept obtained by DL-FUMI is an effective heartbeat prototype and achieves superior performance over comparison algorithms.
\end{abstract}


\section{INTRODUCTION}

{Long-term in home monitoring of vital signs, \eg, blood pressure \cite{kim2015ballistocardiogram, ding2016continuous}, heart rate \cite{zhang2015troika, monkaresi2014machine},  respiratory rate \cite{atalay2015weft} and body temperature \cite{yan2015stretchable}, provides promise for the early treatment of any potential problems, especially for older adults \cite{khan2016monitoring, hristoskova2014ontology}. {According to the statistics reported by} the American Heart Association \cite{benjamin2017heart}, cardiovascular disease (CVD) was the leading primary diagnosis in home health care (18.3\%). Increasingly more and more devices for real-time heart rate monitoring are becoming available.  However, the majority of these devices are intrusive and require continual interaction.  For example, many heart rate monitoring systems require a user to physically wear the system (\eg, as a watch, chest strap, electrodes, {finger sensor, \etc) and/or charge} batteries frequently \cite{bandodkar2014non, banaee2013data}.  In contrast, devices that use ballistocardiography can provide an unintrusive and, thus, relatively low maintenance, comfortable alternative for heart rate monitoring \cite{pinheiro2010theory, hwang2014nocturnal}. These sensing systems record the motion of the human body generated by the sudden ejection of blood into the large vessels at each cardiac cycle \cite{starr1939studies}. Such motion contains rich information and has gained revived interest due to recent development in measurement technology \cite{pinheiro2010theory, inan2015ballistocardiography} and a growing interest in managing chronic health conditions through passive sensors in the home \cite{skubic2015automated, javaid2016elucidating}}.

A hydraulic bed sensor system \cite{rosales2012heartbeat, heise2013non, Su:2017} has been recently developed to collect data for a person during sleep.  This sensor system provides measurement that is the superposition of the ballistocardiogram (BCG) and respiration signals. The purpose of the sensing system is to support continuous, non-intrusive monitoring of vital signs of older adults in an unstructured natural living environment. {The hydraulic bed sensor is placed beneath the mattress and has shown to be flexible, low-cost and non-intrusive for monitoring an individual's heart rate. Compared with other methods such as electrocardiography (ECG), BCG does not need electrodes or clips to be affixed to the patient's body and, thus, is ideal for long term in-home monitoring. However, the lack of saliency and large variability in the BCG signal makes it much more difficult to detect individual heartbeats than with an ECG.} In this work, we develop a supervised multiple instance learning approach to extract an individual's personalized heartbeat signature and, then, use this signature to detect individual heartbeats in the BCG signal and derive the heart rate. 

Supervised machine learning methods typically require that every training data point is individually coupled with an accurate training label. However, in many applications obtaining accurate training labels can be time consuming, expensive, and/or infeasible. Furthermore, annotators may be inconsistent during labeling that leads to inherently imprecise labels. Thus, one often needs to face inaccurately or weakly labeled training data. Also, the majority of supervised machine learning methods assume each data point is a \emph{pure} representative of its associated class. In the case of a BCG system, one avenue for collecting heartbeat groundtruth information is the use of a more intrusive sensor during training (\eg, an ECG or a finger pressure sensor). However, with the use of these, there will be misalignment between the BCG signal and groundtruth signal (ECG or finger sensor). The ECG signal is an electrical signal which starts from SA (sinoatrial) node through the heart to trigger the heartbeat. The BCG signal is a mechanical signal obtained from blood flow, thus, delay between the ECG and BCG is expected. The bed sensor is placed under the mattress which will introduce additional delay. Thus, the exact location in a BCG signal that corresponds to the heartbeat is unknown, making it difficult to align with other sensor signals.  Furthermore, several BCG sensors may be used in parallel to ensure full spatial coverage of a subject.  However, which of these BCG sensors is able to capture a clear heartbeat signal will depend on the subject's position relative to the sensors. Thus, not only is there uncertainty in alignment between groundtruth signals and the BCG, but there is also uncertainty as to which BCG signal from the collection of sensors is most informative and accurate.   Finally, in addition to the label uncertainty, the bed sensor data has components associated with a subject's respiration or body movement. The BCG signals from the bed sensor contains response from not only the heartbeat but also other sources.  Training an accurate classifier or learning a representative target concept from this sort of weakly labeled and mixed data is challenging in practice and requires an approach that can learn from uncertain training labels and mixed data.

Multiple-instance learning (MIL) is a variation on supervised learning for problems with incomplete or uncertain training label information \cite{Dietterich:1997, Maron:1998, Zare:2015fumi}. In MIL, training data are grouped into positive and negative ``bags.'' A bag is defined to be a multi-set of data points where a positive bag contains at least one data point from the target class and negative bags are composed entirely of non-target data. Thus, data point-specific training labels are unavailable.  Given training data in this form, the majority of MIL methods  either: (1) learn target concepts for describing the target class; or (2) train a classifier that can distinguish between individual target and non-target data points and/or bags. Here, \emph{concepts} refer to generalized class prototypes in the feature space; in the case of heartbeat characterization, a concept is the personalized heartbeat pattern of an individual.

Expanding upon the work in \cite{jiao2016heart, jiao2016multipleicpr}, this paper formulates the heartbeat characterization for BCG analysis as a MIL problem.  {However, in contrast to our prior work (which is based upon the earlier $e$FUMI algorithm), this paper presents the DL-FUMI algorithm.   DL-FUMI developed in this paper has the following advantages in comparison to our prior work: (1) DL-FUMI adopts the concept of discriminative dictionary learning  to estimate multiple personalized heartbeat signatures to accounts for the variability in an individual's heartbeat pattern; (2) DL-FUMI introduces the use of a linear mixing model which is more appropriate for BCG signals as opposed to the convex mixing model used in \cite{jiao2016heart}; (3) DL-FUMI applies the Hybrid Detector \cite{Broadwater:2007} for heartbeat detection that can take advantage of the multiple estimated heartbeat signatures; (4) DL-FUMI includes a parameter learning step so the detection parameters (\eg, neighborhood, threshold) are automatically determined from test-on-train procedure; and (5) more comprehensive experimental results (40 subjects) and comparisons are presented as opposed to what we reported previously. }

\section{Related Work}\label{sec:2_literature_review}
Over the past several years, there has been a renewed interest in the use of the BCG.  This revival has been mainly driven by the advancement in piezoelectric sensors, signal processing \cite{pinheiro2010theory, inan2015ballistocardiography, postolache2010physiological} and the need for a noninvasive long-term vital signs monitoring. Many approaches have been developed for analyzing the BCG signal for heartbeat detection and heart rate monitoring. {\cite{kim2016ballistocardiogram} investigates the mechanism for the genesis of BCG waveforms and verifies the proposed model surgically, providing more theoretical foundation for unobtrusive monitoring of heart rate and diagnosis of cardiovascular disease.} \cite{bruser:2011, paalasmaa:2015, rosales2012heartbeat} rely on clustering a collection of BCG signals (or their associated features) to detect heartbeats. Rosales \etal \cite{rosales2012heartbeat} extract three features based on the peak to valley amplitude difference of the BCG signal and then combine the k-means, and Fuzzy C means clustering approaches (CA)  to detect the heartbeat. Heise \etal \cite{Heise2010} use window-peak-to-peak deviation (WPPD)  to estimate the heart rate. WPPD first finds the local maximum at every 0.25 second and then applies a low-pass filter with a cutoff frequency equal to 4Hz to smooth the local maxima. From the smoothed local maxima, WPPD identifies the peak location as the heartbeat position to compute the heart rate. \cite{Lydon2015} first applies a band-pass filter to remove the respiration information and high frequency noise and then computes the short-term energy (EN), from which the peak positions are considered as the heartbeat locations to obtain the heart rate. There are a number of methods that analyze the BCG signal in the frequency domain to extract the heart rate \cite{frequency1, frequency2, frequency3,Su:2012, Su:2017}. Among the frequency domain based approaches, the method by Su \etal \cite{Su:2012, Su:2017} uses the Hilbert transform (HT) to process the data which shows better robustness in the heart rate estimate, and it also has better noise resistance when analyzing the data from the real environment compared to process the data in the time domain. However, the time domain beat-to-beat based approaches are more appealing since they reflect real-time heart rate variability for rhythm pattern analysis \cite{task1996heart}.

\textbf{Multiple-instance learning: } Since the introduction of the MIL framework \cite{Dietterich:1997}, many methods have been proposed and developed in the literature. The majority of MIL approaches \cite{Dietterich:1997, andrews2002support,chen2006miles} focus on learning a classification decision boundary to distinguish between positive and negative instances/bags from the ambiguously labeled data. Although these approaches are effective at training classifiers given imprecise  labels, they generally do not provide an intuitive description or \textit{representative concept} that characterizes the salient and discriminative features of the target class.  The few existing approaches that estimate a target concept include Diverse Density (DD) \cite {Maron:1998} that estimates a concept by minimizing its distance to at least one instance from each positive bag and maximizing its distance from all instances in negative bags. The Expectation-Maximization (EM) version of diverse density (EMDD) \cite{Zhang:2002} iteratively estimates which instance in each positive bag belong to the target class and then only uses those points from the positive bags to estimates a target concept that maximizes the diverse density.  $e$FUMI \cite{Zare:2014whispers, Zare:2015fumi} treats each instance as a convex combination of positive and/or negative concepts and estimates the target and non-target concepts using an EM approach. However, these prototype-based methods only find a single target concept and are, thus, unable to account for any large variation in the target class. To address this, DL-FUMI learns a collection of target dictionary atoms (as well as non-target dictionary atoms) to  be able to characterize any variation in the target class.

\textbf{Sparse Coding and Dictionary Learning: } Sparse coding refers to the task of decomposing a signal into a sparse linear combination of dictionary atoms \cite{mallat1993matching, chen1998atomic, candes2006robust, donoho2006compressed}. Often, in these applications, pre-defined overcomplete dictionaries (\eg, wavelets, curvelets) or a collection of samples from the data are used as the dictionary set.  However, it has been repeatedly shown that, rather than using a pre-defined ``off the shelf'' dictionary, improved signal reconstruction and greater sparsity in the weights can be achieved by estimating a dictionary for a particular application  \cite{aharon2006rm, olshausen1996emergence, olshausen1997sparse}. The unsupervised dictionary learning approaches generally estimate the dictionary and a corresponding set of sparse weights by minimizing the reconstruction error between the original data and its representation using the estimated dictionary and weights. For example, K-SVD \cite{aharon2006rm} alternates between a sparse coding step using orthogonal matching pursuit (OMP) \cite{pati1993orthogonal, tropp2007signal} and a dictionary learning step using singular value decomposition until a preset reconstruction error is reached. However, recent approaches \cite{zhang2010discriminative, jiang2013label, mairal2012task} demonstrate that classification performance can be improved via supervised dictionary learning.  

The proposed DL-FUMI shares the goal of aforementioned approaches to learn an effective set of dictionary atoms for the target class but successfully handles problems in which only imprecise multiple instance learning type labels are available. The model assumed by DL-FUMI, in which each instance is a linear combination of target and non-target dictionary atoms, provides the ability to both discriminate between target and non-target characteristics, account for variation in the target class, and learn a dictionary that provides effective representative prototypes of target and non-target instances.  Furthermore, DL-FUMI is unique in that each instance (both positive and negative) shares a common set of non-target/background dictionary atoms.   In this way, the estimated target dictionary is discriminative and accounts only for the unique variation in positive class since it is estimated from the residual information not accounted for by the shared non-target/background dictionary.

\section{HYDRAULIC BED SENSOR}\label{sec:3_sensor}

The hydraulic bed sensor (HBS) developed at the Center for Eldercare and Rehabilitation Technology (CERT) at the University of Missouri is a BCG device providing a low-cost, noninvasive and robust solution for capturing physiological parameters during sleep \cite{rosales2012heartbeat, heise2013non, Su:2017}. The HBS was designed to maintain an imperceptible flat profile and to be used beneath a bed mattress. The system is comfortable for subjects lying on the mattress (\ie, noninvasive), easy to install, watertight, and durable.

\begin{figure}
	\begin{center}
		\subfigure[Sensor and Embedded System]{
			\includegraphics[width=8cm]{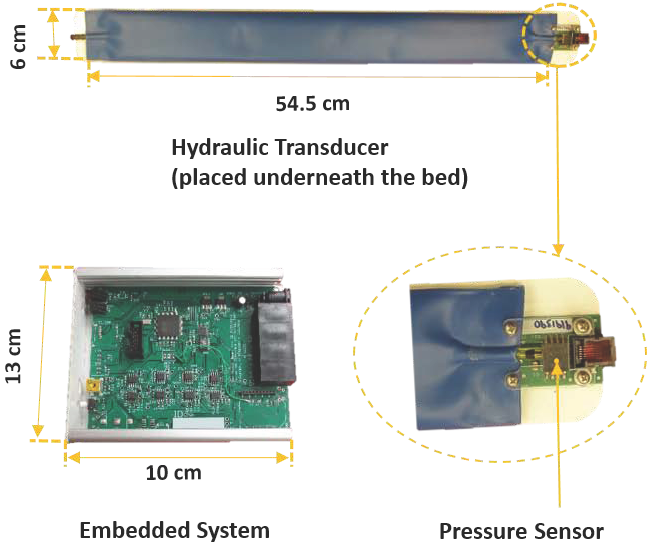} \label{fig:bed_system}}
		\subfigure[Transducer placement]{
			\includegraphics[width=7.5cm]{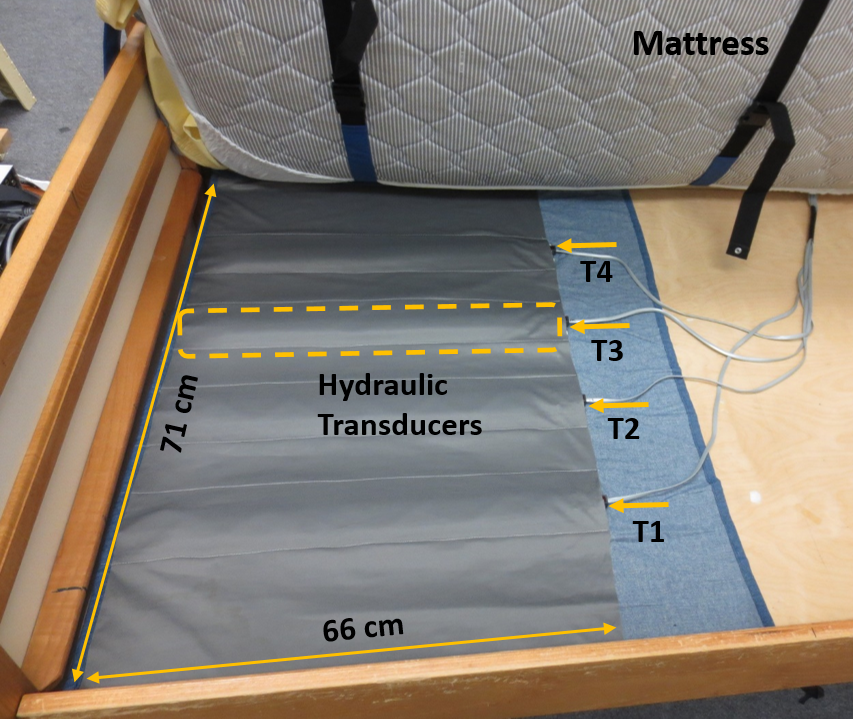} \label{fig:bed_transducer}}
		\caption{Hydraulic Bed Sensor System. (a) Hydraulic transducer filled with 0.4 liter of water (top) and embedded system that collects and transmits data from four transducers (bottom). (b) Transducer placement: the transducers are separated 14cm from each other to provide enough coverage for human body.}\label{fig:bed_sensor}
	\end{center}
\vspace{-3mm}
\end{figure}

The HBS is composed of a transducer and a pressure sensor as shown in Fig. \ref{fig:bed_system}. The transducer was designed to be placed under the subject's upper torso. It is 54.5 cm long, 6 cm wide, and is filled with 0.4 liters of water \cite{rosales2012heartbeat, heise2013non, Su:2017}. The integrated silicon pressure sensor (Freescale MPX5010GP) attached to the end of the transducer is used for measuring the vibration of human body arising from each heartbeat. It captures the information of heartbeat together with respiration and motion disturbance. The signal from each transducer is then amplified, filtered and sampled at 100 Hz.

In order to ensure enough coverage, four transducers are placed in parallel underneath a mattress as shown in Fig. \ref{fig:bed_transducer}. The four transducers are identical and independent, but the data quality collected by those four transducers could vary depending on the sleeping position, type of mattress (\eg, material, thickness, \etc) and the physical characteristics of the subject (\eg, age, body mass index (BMI), \etc).

\section{DATA DESCRIPTION}\label{sec:data_desc}

Given the hydraulic bed sensor system described above, in this study, we use the dataset collected from 40 subjects at the CERT at the University of Missouri. The data collection from human subjects has been approved by the Institutional Review Board (IRB) at the University of Missouri. To prepare for the data collection, each subject was asked to lie flat on their back for 10 minutes. Then the subjects were asked to perform a bicycling for 3 minutes and the BCG data was recorded again on the supine position for at least 5 minutes. The gender, age, weight and height of the subjects are listed in Table \ref{tab:heart_rate_error}. There were 7 females and 33 males. Their ages, weights, heights and BMIs are 18 - 49, 48 - 127 (Kg), 156 - 190 (cm), 18.3 - 37.9, with the average as 29.2, 76.9 (Kg), 175.2 (cm) and 24.9, respectively.  The BCG signal was sampled at 100Hz and filtered by a six order Butterworth band-pass filter with 3dB cutoff frequency at 0.4Hz and 10Hz to remove the respiratory component and high frequency noise. 

	  \begin{figure}
	  	\begin{center}
	  		\includegraphics[width=9.3cm]{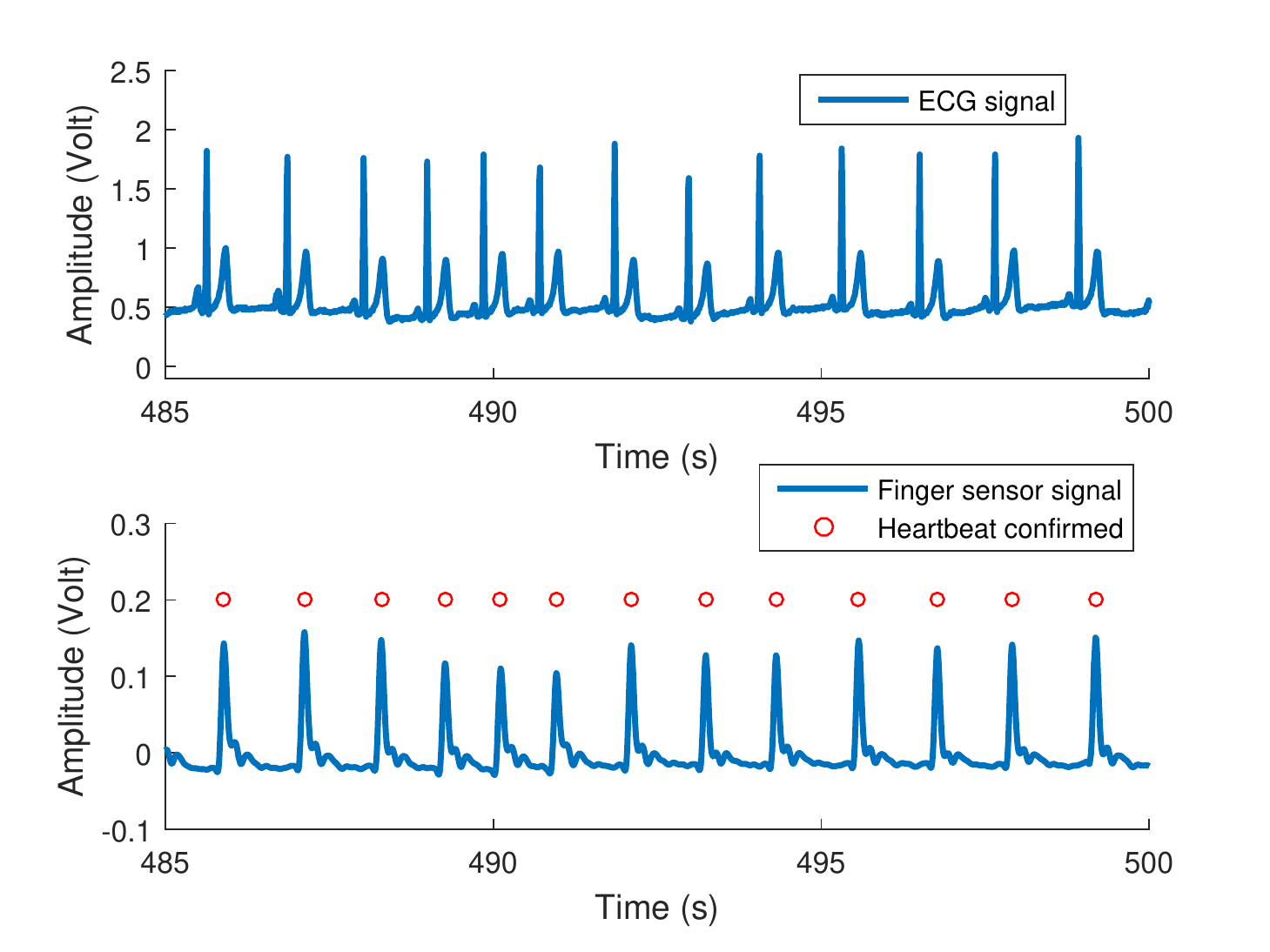}
	  		\caption{Comparison of GT by finger sensor and ECG. The  signal received by finger sensor has some amount of delay, but the beat to beat intervals collected by the finger sensor match the ECG well.}\label{fig:fingerseonsro_ECG_plot}
	  	\end{center}
	  \end{figure}

{For ground truthing, a piezo-electric based pressure transducer from ADInstruments (TN1012/ST \cite{finger_sensor}) was hooked on the subject's right-hand index finger. The piezo-electric pulse transducer records the expansion and contraction force applied to the active surface of the transducer ejected by the changes from blood pressure, converts the force into an electrical analog signal which can then be used to determine heart rate. The subjects were asked to stay still during collection. }
	
{The heart rate ground truth is computed by averaging the beat-to-beat heart rate confirmed by finger sensor within a one minute sliding window. Since the peak of the finger pulse could be affected by the well-known arterial wave reflection \cite{mukkamala2015toward}, we provide a comparison of the ground truth acquired by finger sensor and ECG as a verification.  Fig. \ref{fig:fingerseonsro_ECG_plot} shows a comparison between the GT acquired by ECG and finger sensor, respectively. Although the signal received by finger sensor has some amount of delay relative to ECG, the beat to beat intervals collected by the finger sensor match those from the ECG very well. Also it has been found that the arterial wave reflection is more clearly observed from  older people \cite{swillens2008assessment} and the subjects discussed in this paper are mostly younger healthy people which are much less affected by the arterial wave reflection. Moreover, in contrast to the electrocardiogram, the signal captured by the hydraulic bed sensor is BCG resulted from the mechanical cardiogenic movements of the body. We believe that in this situation, the finger sensor which is not based on the electrical signal can serve as a more suitable ground truth measurement than the electrocardiogram.}

\begin{figure}
	\begin{center}
		\includegraphics[width=8.9cm]{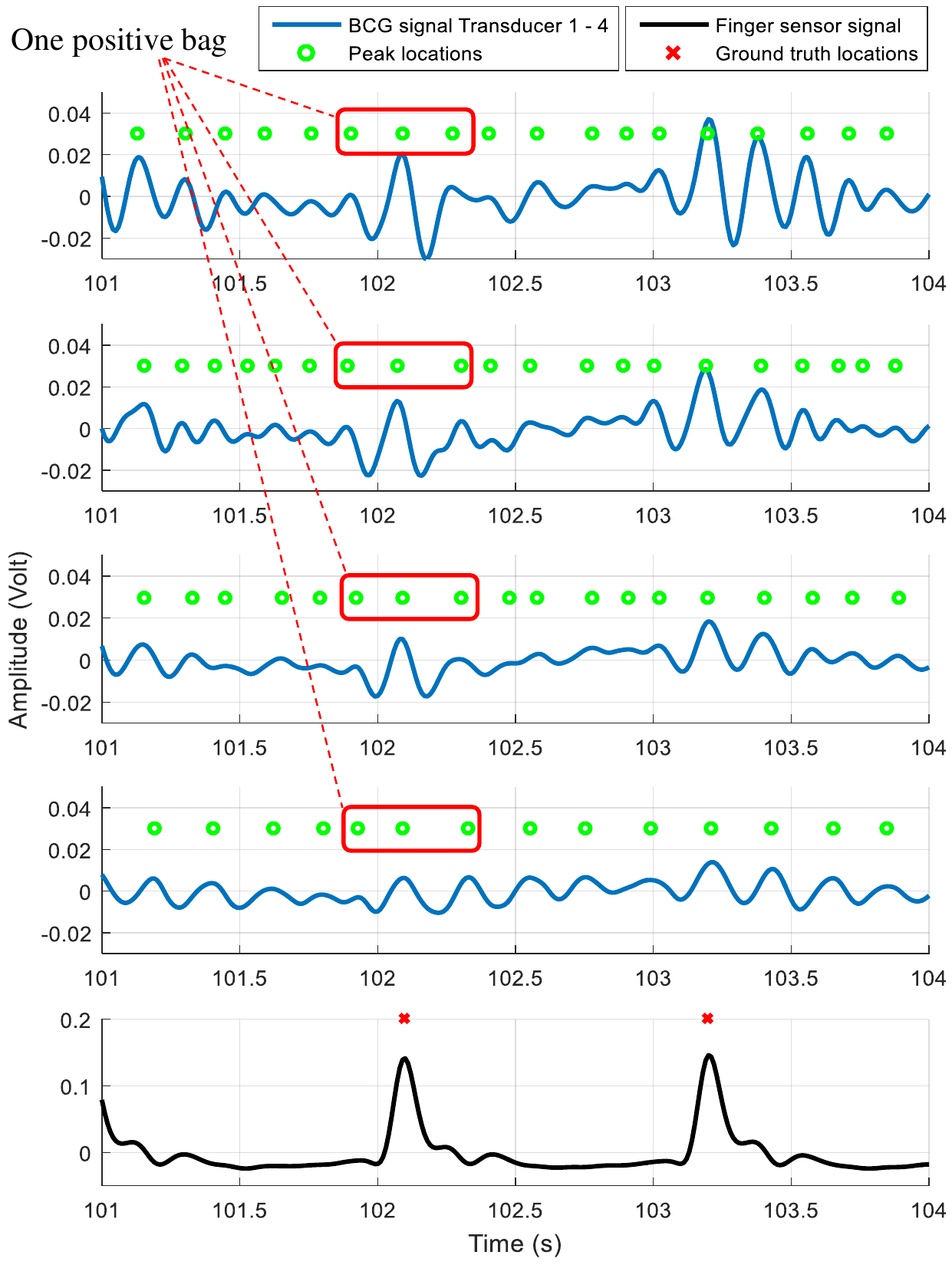}
		\caption{BCG Signal and Ground Truth Plot. Peaks location were first found as candidate heartbeat (J-peak) locations (green circles), then 3 instances from each transducer closest to the ground-truth location marked by the finger sensor (red cross) were grouped together as a positive bag; instances between two positive bags not placed in any positive bag were grouped together to form one negative bag.}\label{fig:BCG_FS_plot}
	\end{center}
\vspace{-6mm}
\end{figure}

\textbf{Feature Extraction from Ballistocardiograms Time Series: } In this paper, we used the filtered time domain segments from received BCG signals as training features. Specifically, for each subject, we found the signal peaks for the entire BCG signal received by four transducers as candidate J-peak locations (possible heartbeat locations). For each peak, we extract a data segment (simply called instance) that is 91 samples long and centered at the peak (corresponding to 0.91s signal, 45 samples before and after the peak) as the training feature for this peak location. The feature length (0.91s, at 100 Hz sampling rate) was determined empirically and found to be the typical length of a heartbeat pattern. Fig. \ref{fig:BCG_FS_plot} shows example filtered BCG signals collected by four transducers (blue plots) and the corresponding finger sensor ground truth information (black plots), where the green circles denote every peak location of the filtered BCG signal.

\section{Multiple Instance Learning Problem in Ballistocardiograms}
From Fig. \ref{fig:BCG_FS_plot}, it can be seen that near the ground truth locations denoted by the finger sensor, there are prominent peak patterns measured by the four BCG transducers corresponding to heartbeats. However, although all of the sensors are expected to be capturing each corresponding heartbeat signal simultaneously, there is unavoidable misalignment between the finger sensor and each of the BCG pressure sensors. Furthermore, depending on the location and position of the subject lying on the bed, which of these BCG sensors are able to capture a clear heartbeat signal is difficult to determine. These multiple labeling uncertainties in the training data cast more difficulties to traditional supervised learning methods for heartbeat detection and heart rate estimation from a BCG signal.

\begin{figure}
	\begin{center}
		\includegraphics[width=9cm]{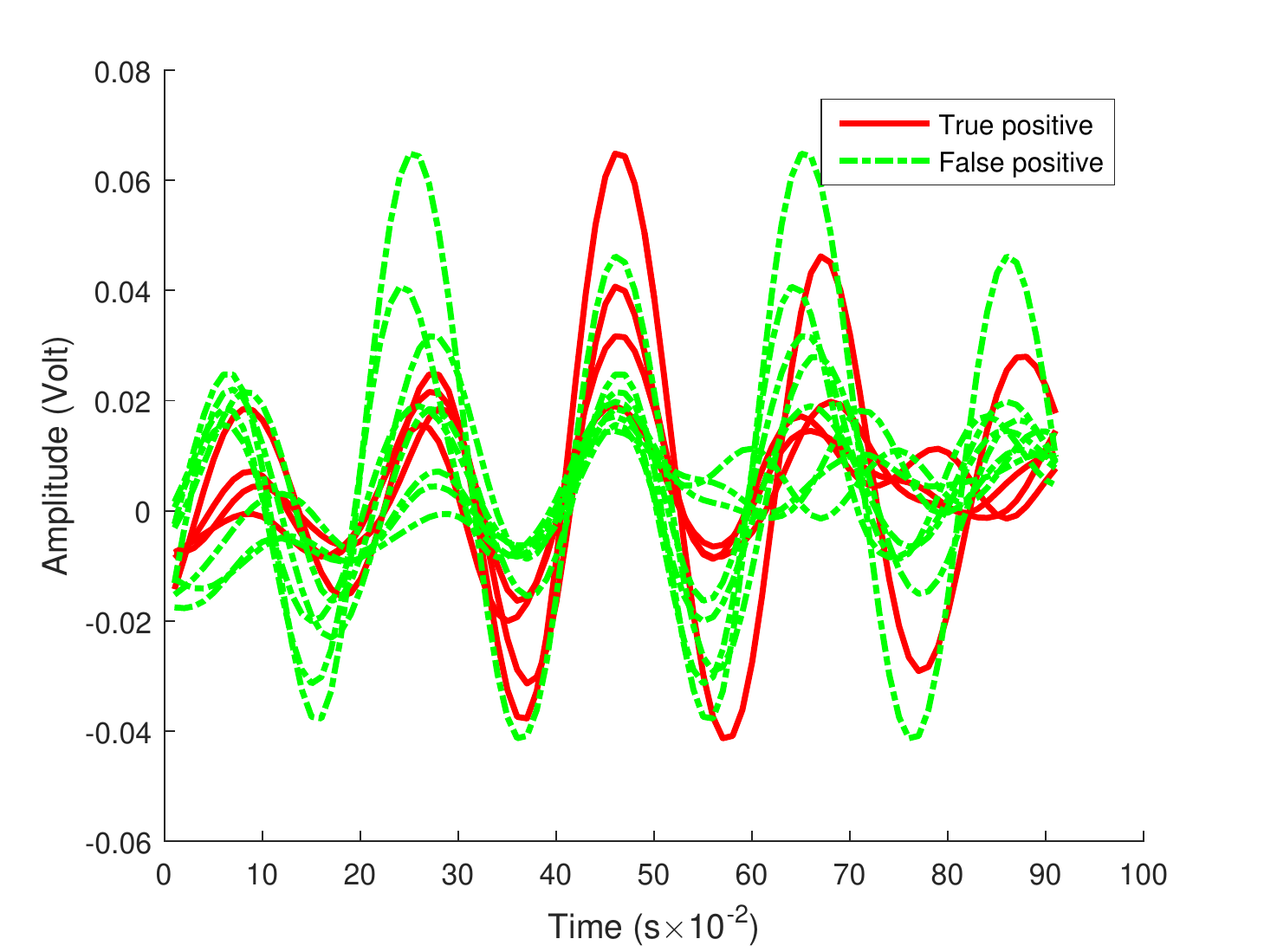}
		\caption{Plot of one positive bag. A positive bag is also mixed with both true positive instances (heartbeat signal, in red) and false positive instances (non-heartbeat signal, in dotted green), where the true positive heartbeat signals tend to have more prominent J-peaks.}\label{fig:pos_bag_plot}
	\end{center}
\end{figure}

In this paper, we continue to investigate the idea of training ``bags'' to address label uncertainty as well as mis-collection of heartbeat signals in the BCG data. Specifically, each of the extracted sub-signals is treated as individual data points (or ``instances'') during training. Each positive labeled training bag (shown as red rectangles in Fig. \ref{fig:BCG_FS_plot}) was formed by grouping the 12 instances across the four transducers (3 instances for each transducer) that are close in time to the ground-truth location marked by the finger sensor (shown as a red cross in Fig. \ref{fig:BCG_FS_plot}).  Similarly, one negative bag was formed by grouping instances from four transducers between two positive bags that were not included in any positive bag. Fig. \ref{fig:pos_bag_plot} shows an example positive bag.  In this figure, we can see that a positively labeled bag contains both true heartbeat patterns (shown in red) and non-heartbeat patterns ( shown in dotted green). From Fig. \ref{fig:pos_bag_plot} it can also be seen that the assumed heartbeat patterns tend to have more prominent J-peaks. The proposed DL-FUMI algorithm is expected to learn a set of discriminative subject-specific heartbeat concepts from  training bags of this type. After learning the heartbeat concept, a signature based detector can then be applied for real-time heartbeat monitoring and heart rate estimation.

\section{DL-FUMI HEARTBEAT CONCEPT LEARNING Algorithm}\label{sec:4_DLFUMI}
\begin{figure*}[!t]
	\begin{equation}
	F=\frac{1}{2}\sum_{i=1}^Nw_i\bigg\|(\mathbf{x}_i-z_i\sum_{t=1}^T\alpha_{it}\mathbf{d}_t^+-\sum_{k=1}^M\alpha_{ik}\mathbf{d}_k^-)\bigg\|_2^2+\lambda\sum_{i=1}^{N}w_i\left\|\begin{bmatrix}z_i \boldsymbol{\alpha}^+_i \\ \boldsymbol{\alpha}^-_i \end{bmatrix} \right\|_1+\sum_{k=1}^M\sum_{t=1}^{T}\gamma_{kt}\langle\mathbf{d}_k^-,\mathbf{d}_{t^{\text{old}}}^+\rangle
	\label{eqn:gFUMI}\tag{3}
	\end{equation}

	\begin{small}
		\begin{equation}
		E[F]=\sum_{\substack{z_i\in\{0,1\}}} P(z_i|\mathbf{x}_i, \boldsymbol{\theta}^{(l-1)})  \left[ \frac{1}{2}\sum_{i=1}^N w_i \left\| \mathbf{x}_i - z_i\sum_{t=1}^T\alpha_{it}\mathbf{d}_t^+ - \sum_{k=1}^M\alpha_{ik}\mathbf{d}_k^-\right\|_2^2+\lambda\sum_{i=1}^{N}w_i\left\|\begin{bmatrix} z_i \boldsymbol{\alpha}^+_i \\ \boldsymbol{\alpha}^-_i \end{bmatrix} \right\|_1\right]+\sum_{k=1}^M\sum_{t=1}^{T}\gamma_{kt}\langle\mathbf{d}_k^-,\mathbf{d}_{t^{\text{old}}}^+\rangle
		\label{eqn:E_gFUMI}\tag{4}
		\end{equation}
		\hrulefill
	\end{small}
\end{figure*}

{Let $\mathbf{X}=\left[\mathbf{x}_1,\cdots,\mathbf{x}_N\right]\in\mathbb{R}^{d\times N}$ be training data where $d$ is the dimensionality of an instance, $\mathbf{x}_i$, and $N$ is the total number of training instances. The data is grouped into $K$ \textit{bags},  $\mathbf{B} = \left\{ \mathbf{B}_1, \ldots, \mathbf{B}_K\right\}$, with associated binary bag-level labels, $L = \left\{L_1, \ldots, L_K\right\}$ where $L_j \in \left\{ 0, 1\right\}$ and $\mathbf{x}_{ji} \in \mathbf{B}_j$ denotes the $i^{th}$ instance in bag $\mathbf{B}_j$. Given training data in this form, DL-FUMI models each instance as a sparse linear combination of target and/or background atoms $\mathbf{D}$, $\mathbf{x}_i\approx\mathbf{D}\boldsymbol{\alpha}_i$, where $\boldsymbol{\alpha}_i$ is the sparse vector of  weights for instance $i$. Positive bags (\ie, $\mathbf{B}_j$ with $L_j = 1$, denoted as $\mathbf{B}_j^+$) contain at least one instance composed of some target:}

\begin{eqnarray}
&&\text{if }L_j = 1,  \nonumber \exists \mathbf{x}_i \in \mathbf{B}_j^+ \text{ s.t. } \\
&&\mathbf{x}_i = \sum_{t=1}^{T}\alpha_{it}\mathbf{d}_t^+ + \sum_{k=1}^{M} \alpha_{ik}\mathbf{d}_{k}^-+\boldsymbol{\varepsilon}_{i}, \alpha_{it} \ne 0,
\label{eq:l1}
\end{eqnarray}
{where $\boldsymbol{\varepsilon}_i$ is a noise term.   However, the number of instances in a positive bag with a target component is unknown.}

If $\mathbf{B}_j$ is a negative bag (\ie, $L_j = 0$, denoted as $\mathbf{B}_j^-$), then this indicates that $\mathbf{B}_j^-$ does not contain any target:
\begin{equation}
\text{if }L_j = 0,  \forall \mathbf{x}_i \in \mathbf{B}_j^-, \mathbf{x}_i =  \sum_{k=1}^{M} \alpha_{ik}\mathbf{d}_{k}^-+\boldsymbol{\varepsilon}_{i}
\label{eq:l2}
\end{equation}

{Given this problem formulation, the goal of DL-FUMI is to estimate the dictionary\footnotemark[1] $\mathbf{D}=\begin{bmatrix}\mathbf{D}^+ & \mathbf{D}^-\end{bmatrix}\in\mathbb{R}^{d\times (T+M)}$, where $\mathbf{D}^+ = \left[\mathbf{d}_{1}^+,\cdots,\mathbf{d}_{T}^+\right]$ are the $T$ target  atoms and $\mathbf{D}^- = \left[\mathbf{d}_{1}^-,\cdots,\mathbf{d}_{M}^-\right]$ are the $M$ background  atoms. This is accomplished by minimizing \eqref{eqn:gFUMI} which is proportional to the {complete} negative data log-likelihood, where $\boldsymbol{\alpha}_{i}^{+}$ and $\boldsymbol{\alpha}_{i}^{-}$ are subsets of $\boldsymbol{\alpha}_{i}$ corresponding to $\mathbf{D}^+$ and $\mathbf{D}^-$, respectively.
This model (as in \eqref{eq:l1} and \eqref{eq:l2}), differs significantly from most class-specific dictionary learning methods since the background dictionary set is shared among both classes.  This provides the ability to estimate a target dictionary $\mathbf{D^+}$ that is very distinct from non-target data and characterizes the salient and discriminative features of the target class.}

{The first term in \eqref{eqn:gFUMI} computes the squared residual error between each instance and its estimate using the dictionary.  In this term, a set of hidden binary latent variables $\left\{z_i\right\}_{i=1}^{N}$ that indicate whether an instance is or is not a target (\ie, $z_i = 1$ when $\mathbf{x}_i$ contains target) are introduced.  For all points in negative bags, $z_i = 0$.  For points in positive bags, the value of $z_i$ is unknown.   Also, a weight $w_i$ is included where $w_i= 1$ if $\mathbf{x}_i \in \mathbf{B}_j^-$ and $w_i = \psi$ if $\mathbf{x}_i \in \mathbf{B}_j^+$ where $\psi$ is a fixed parameter.  This weight helps balance terms when there is a large imbalance between the number of negative and positive instances.}
\footnotetext[1]{$\begin{bmatrix} \mathbf{A} & \mathbf{B}\end{bmatrix}$ and $\begin{bmatrix} \mathbf{A} \\ \mathbf{B}\end{bmatrix}$ are the concatenation of arrays $\mathbf{A}$ and $\mathbf{B}$ horizontally and vertically, respectively.}

{The second term is an $l_1$ regularization term to promote sparse weights. It also includes the latent variables, $z_i$, to account for the uncertain presence of target in positive bags.}

{The third term is a {robust penalty term} that promotes discriminative target atoms (and inspired by a term presented in \cite{ramirez2010classification}). Instead of using a fixed penalty coefficient, we introduce an adaptive coefficient $\gamma_{kt}$ defined as:}
\setcounter{equation}{4}
\begin{equation}
\gamma_{kt}=\Gamma\frac{\langle\mathbf{d}_{k}^-,\mathbf{d}_{t}^+\rangle}{\|\mathbf{d}_{k}^-\|\|\mathbf{d}_{t}^+\|}=\Gamma\cos\theta_{kt},
\label{eqn:robust_t}
\end{equation}%
{\textcolor{mygreen}{}where $\theta_{kt}$ is the vector angle between the $k^{th}$ background atom and the $t^{th}$ target  atom. Since $sign(\gamma_{kt}) = sign(\langle\mathbf{d}_{k}^-,\mathbf{d}_{t}^+\rangle)$, this discriminative term is always positive and will add large penalty when $\mathbf{d}_k^-$ and $\mathbf{d}_t^+$ have similar shape.  Thus, this term encourages a discriminative dictionary by promoting background atoms that are orthogonal to  target atoms. In implementation, $\gamma_{kt}$ is updated once per iteration using $\mathbf{d}_{k^{old}}^-$ and $\mathbf{d}_{t^{old}}^+$ which are the dictionary values from the previous iteration.}

\begin{algorithm}
	\caption{DL-FUMI EM algorithm}
	\algsetup{indent=2em}
	\begin{algorithmic}[1]
		\STATE Initialize $\boldsymbol{\theta}^{0} = \left\{ \mathbf{D}, \left\{ \boldsymbol{\alpha}_i \right\}_{i=1}^N  \right\}$, $l = 1$
		\REPEAT
		\STATE \textbf{\emph{E-step}}: Compute  $P(z_i|\mathbf{x}_i, \boldsymbol{\theta}^{(l-1)})$
		\STATE \textbf{\emph{M-step}}:
		\STATE Update $\mathbf{d}_t^+$ using \eqref{eqn:update_dt}, $\mathbf{d}_t^+\gets\frac{1}{\|\mathbf{d}_t^+\|_2}\mathbf{d}_t^+, t=1,\cdots,T$
		\STATE Update $\mathbf{d}_k^-$ using \eqref{eqn:update_dk}, $\mathbf{d}_k^-\gets\frac{1}{\|\mathbf{d}_k^-\|_2}\mathbf{d}_k^-, k=1,\cdots,M$
		\FOR{$q \gets 1$ to $iter$}
		\STATE  \hspace{-2mm} Update $\left\{ \boldsymbol{\alpha}_i \right\}_{i=1}^{N^+}$ for $\mathbf{x}_i \in \mathbf{B}_j^+$ using \eqref{eqn:alpha_plus_update}, \eqref{eqn:sf_alpha_pos}\\
		\STATE \hspace{-2mm} Update $\left\{ \boldsymbol{\alpha}_i \right\}_{i=1}^{N^-}$ for $\mathbf{x}_i \in \mathbf{B}_j^-$ using \eqref{eqn:alpha_minus_update}
		\ENDFOR
		\STATE $l \gets l + 1$
		\UNTIL{Stopping criterion met}
		\RETURN $\mathbf{D}$, $\left\{ \boldsymbol{\alpha}_i \right\}_{i=1}^N$\\
		
	\end{algorithmic}
	\label{alg:gFUMI}
\end{algorithm}

\section{DL-FUMI Optimization} \label{sec:5_optimization}

{Expectation-Maximization is used to optimize \eqref{eqn:gFUMI} and estimate $\mathbf{D}$.  During optimization, the fact that many of the binary latent variables $\left\{z_i\right\}_{i=1}^N$ are unknown is addressed by taking the expected value of the log likelihood with respect to $z_i$ as shown in \eqref{eqn:E_gFUMI}.  In  \eqref{eqn:E_gFUMI}, $\boldsymbol{\theta}^{l} = \left\{ \mathbf{D}, \left\{ \boldsymbol{\alpha}_i \right\}_{i=1}^N \right\}$is the set of parameters estimated at iteration $l$ and  $P(z_i|\mathbf{x}_i, \boldsymbol{\theta}^{(l-1)})$ is the probability that each instance is or is not a true target instance. During the E-step of each iteration, $P(z_i|\boldsymbol{x}_i, \boldsymbol{\theta}^{(l-1)})$ is computed as:}
\begin{eqnarray}
& &P(z_i|\mathbf{x}_i, \boldsymbol{\theta}^{(l-1)}) = \nonumber  \\
& &\left\{ \begin{array}{l l}
e^{-\beta \left\| \mathbf{x}_i - \sum_{k=1}^M\alpha_{ik}\mathbf{d}_k^-\right\|_2^2} & \text{if } z_i = 0, L_j = 1\\
1-e^{-\beta \left\| \mathbf{x}_i - \sum_{k=1}^M\alpha_{ik}\mathbf{d}_k^-\right\|_2^2} & \text{if } z_i = 1, L_j = 1\\
0 &  \text{if } z_i = 1, L_j = 0\\
1 &  \text{if } z_i = 0, L_j = 0\\
\end{array}\right.,
\label{eqn:pz1}
\end{eqnarray}    {where $\beta$ is a fixed scaling parameter. If $\mathbf{x}_i$ is a non-target instance, then it should be characterized by the background atoms well, thus $P(z_i=0|\mathbf{x}_i, \boldsymbol{\theta}^{(l-1)}) \approx 1$. Otherwise, if $\mathbf{x}_i$ is a true target instance, it will not be characterized well using only the background atoms and $P(z_i=1|\mathbf{x}_i, \boldsymbol{\theta}^{(l-1)}) \approx 1$. Please note,  $z_i$ is unknown only for the positive bags; for the negative bags, $z_i$ is fixed to 0.  This constitutes the \emph{E-step} of the EM algorithm.}

{The \emph{M-step} is performed by iteratively optimizing \eqref{eqn:E_gFUMI} for each of the desired parameters. The dictionary $\mathbf{D}$ is updated atom-by-atom using a block coordinate descent scheme \cite{bertsekas1999nonlinear, mairal2010online}.  The sparse weights, $\left\{ \boldsymbol{\alpha}_i \right\}_{i=1}^N$, are updated using an iterative shrinkage-thresholding algorithm \cite{figueiredo2003algorithm, daubechies2003iterative}.  For readability, the derivation of update equations are described in Sec. \ref{sec:gFUMI_update}.  The method is summarized in Alg. \ref{alg:gFUMI}.}

\section{Heartbeat Detection using Estimated Dictionary} \label{sec:6_classif}

Once the heartbeat concept set $\mathbf{D}$ has been estimated using DL-FUMI,  heartbeat detection on test data can be performed using the hybrid structured detector (HSD) \cite{broadwater2004hybrid, Broadwater:2007}. The hypotheses used for HSD are:
\begin{eqnarray}
&&\mathbf{H}_0: \mathbf{x}\sim\mathcal{N}\left(\mathbf{D}^-\boldsymbol{\alpha}^-, \sigma_0^2\boldsymbol{\Sigma} \right)\nonumber \\
&&\mathbf{H}_1: \mathbf{x}\sim\mathcal{N}\left(\mathbf{D}\boldsymbol{\alpha}, \sigma_1^2\boldsymbol{\Sigma}\right)
\label{eqn:HSD_model}
\end{eqnarray}
where $\boldsymbol{\Sigma}$ is the training background covariance and $\boldsymbol{\alpha}$ and $\boldsymbol{\alpha}^-$ are the corresponding sparse codes of test data $\mathbf{x}$ given full concept set $\mathbf{D}$ and background concept set $\mathbf{D}^-$, respectively.

The confidence that the $i^{th}$ signal is target can be computed using the generalized likelihood ratio test (GLRT) of hypotheses \eqref{eqn:HSD_model},
  \begin{equation}
  \Lambda_{HSD}(\mathbf{x}_i,\mathbf{D})=\frac{(\mathbf{x}_i-\mathbf{D}^-\boldsymbol{\alpha}^-_i)^T\boldsymbol{\Sigma}^{-1}(\mathbf{x}_i-\mathbf{D}^-\boldsymbol{\alpha}^-_i)}{(\mathbf{x}_i-\mathbf{D}\boldsymbol{\alpha}_i)^T\boldsymbol{\Sigma}^{-1}(\mathbf{x}_i-\mathbf{D}\boldsymbol{\alpha}_i)},
  \label{eqn:HSD}
  \end{equation}

Eq. \eqref{eqn:HSD} indicates that if the reconstruction error of signal $\mathbf{x}_i$ using only the background concept (the numerator) is much larger than that using the entire concept, then the target concept is needed in the reconstruction of signal $\mathbf{x}_i$ and $\mathbf{x}_i$ is likely to be a target.  The HSD explicitly models the mixture in BCG signals, utilizes the multiple target concepts and provides a sub-signal detection alternative.

\section{EXPERIMENTAL RESULTS}\label{sec:7_results}
{In this section, three experiments were conducted to show the effectiveness and robustness of the proposed DL-FUMI algorithm. In the first experiment, for each subject, DL-FUMI was applied to the first five minutes heartbeat signal to train a personalized heartbeat concept. It is applied to the next 5 minutes to perform real time signature based heartbeat detection. In order to perform heartbeat detection without training for each subject individually, in the second experiment, a batch training on a group of subjects was conducted to obtain several heartbeat concepts, which are tested on the rest of the subjects. In the third experiment, heartbeat concepts estimated individually (for every subject at rest) from the first experiment was applied to five minutes of data after bicycling exercise to evaluate if signatures learned are applicable to changes in heart rate. }

\subsection{Heartbeat Detection and Rate Estimation from Individually Trained Heartbeat Concept}\label{sec:exp_indivi_training}

\begin{figure}
	\begin{center}
		\includegraphics[width=8cm]{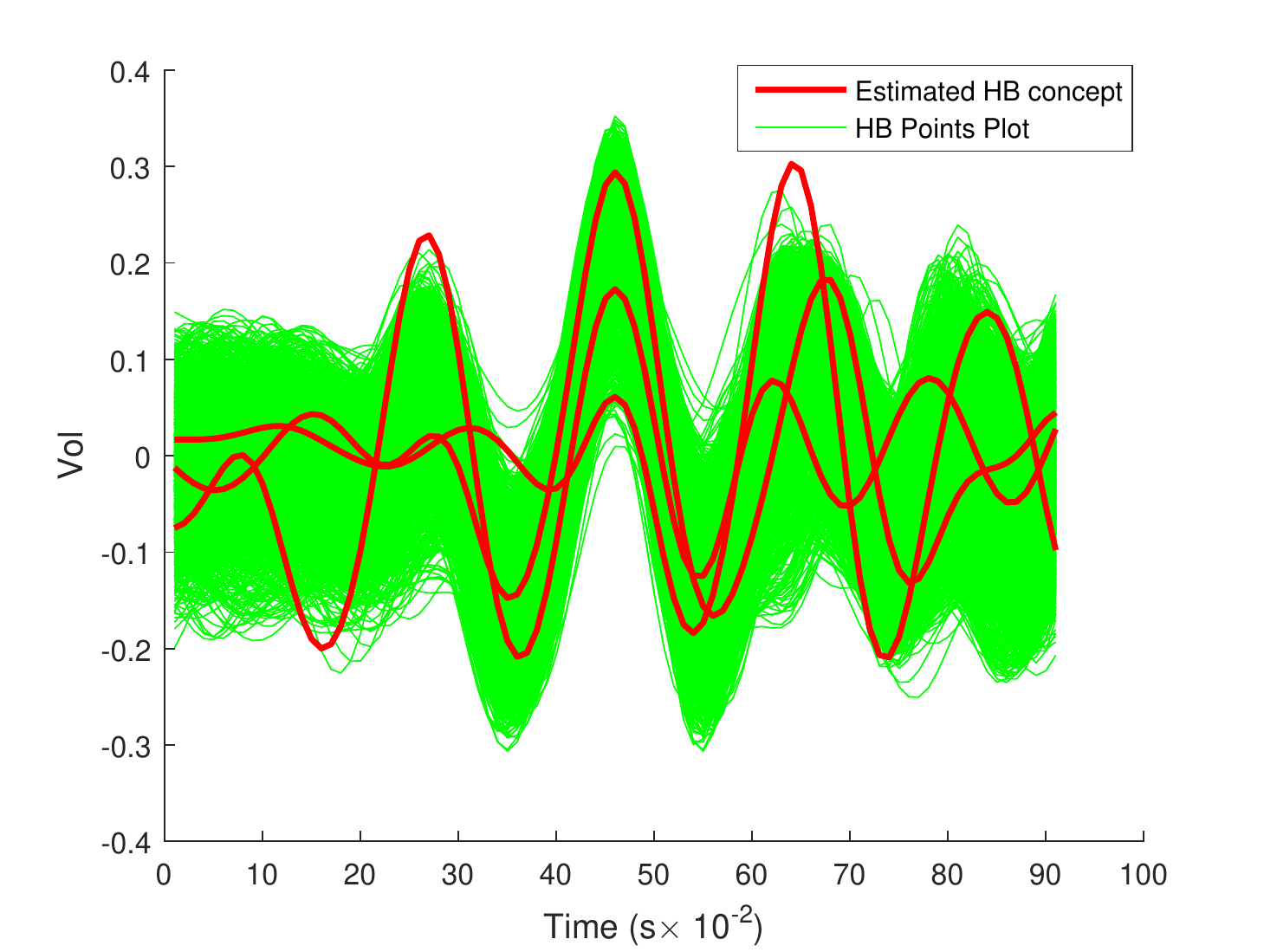}
		\caption{Estimated Heartbeat Concept by DL-FUMI. The multiple heartbeat concepts estimated by DL-FUMI better account for the variability in heartbeat signals.}\label{fig:HB_concepts_DLFUMI}
	\end{center}
\end{figure}

\begin{figure}
	\begin{center}
		\includegraphics[width=8cm]{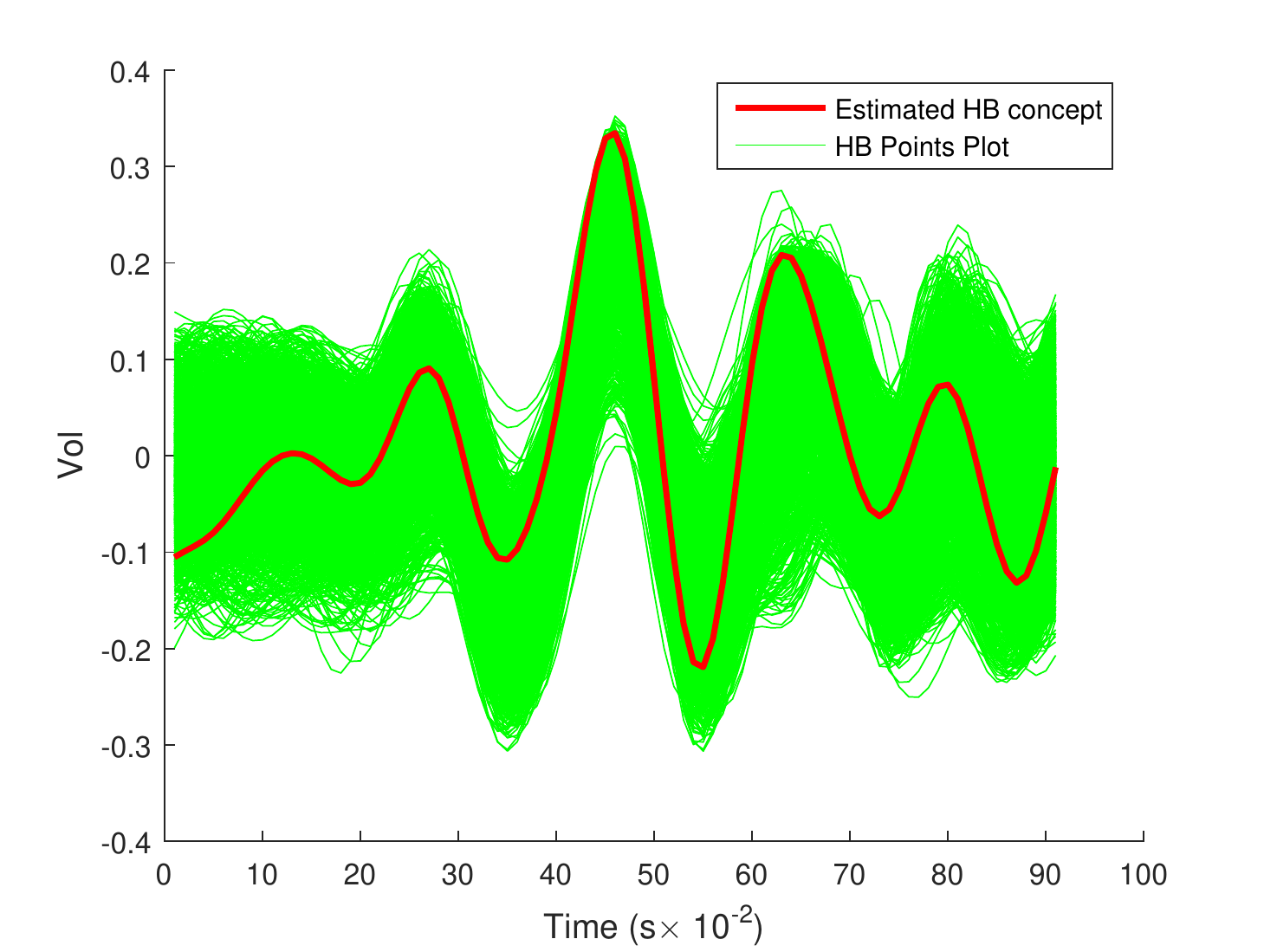}
		\caption{Estimated Heartbeat Concept by EMDD. A single heartbeat concept estimated by EMDD fits the plots of the heartbeat signal well, but fails to maintain the variability in the heartbeat prototype.}\label{fig:HB_concepts_emdd}
	\end{center}
\end{figure}

\begin{figure}
	\begin{center}
		\includegraphics[width=8cm]{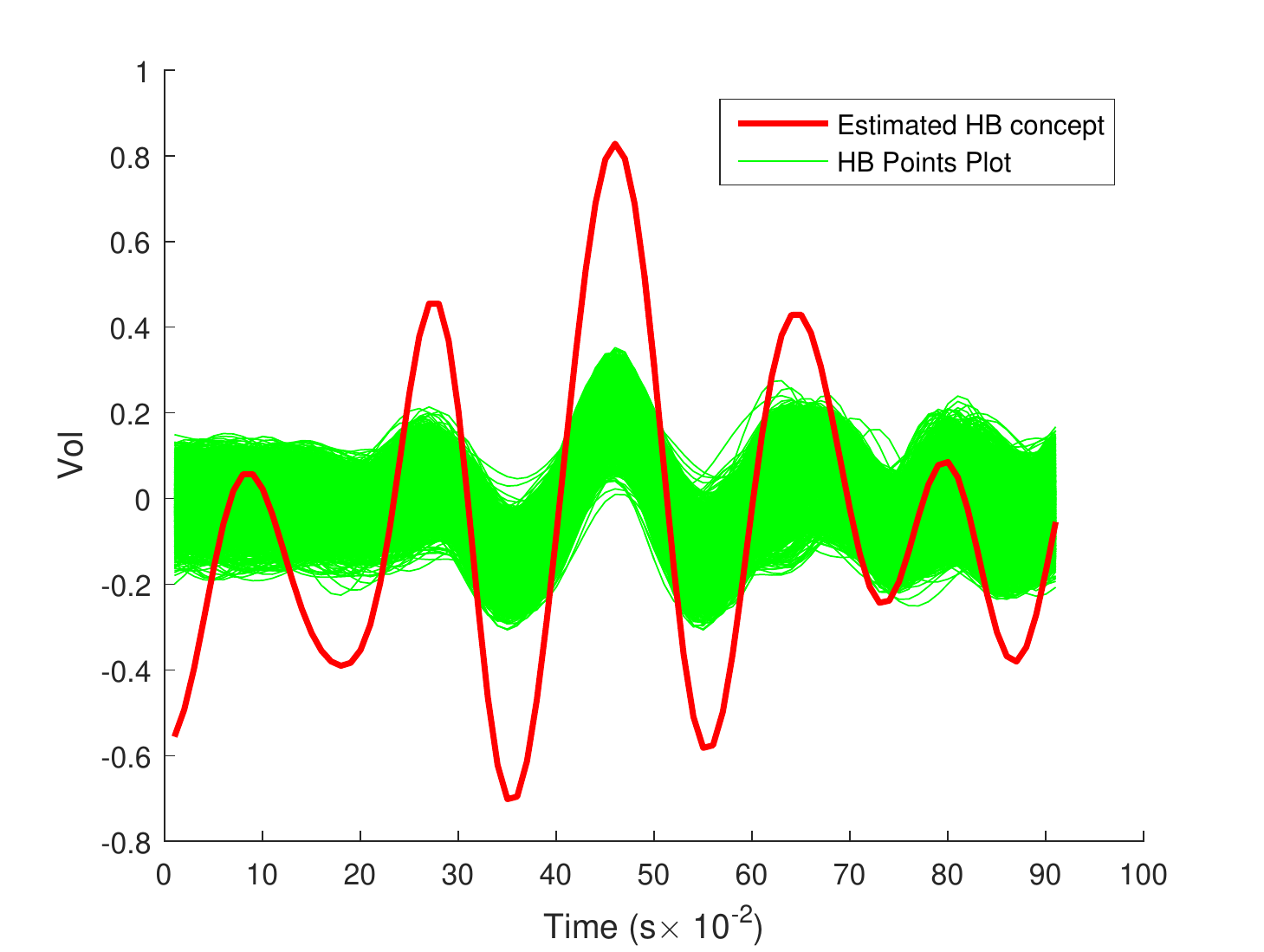}
		\caption{{Estimated Heartbeat Concept by $e$FUMI. The heartbeat concept estimated by $e$FUMI from the convex model is the vertex of the assumed convex hall and is not a fidelity of one's personalized heartbeat concept.} }\label{fig:HB_concepts_eFUMI}
	\end{center}
\end{figure}

\begin{figure}
	\begin{center}
		\includegraphics[width=9.3cm]{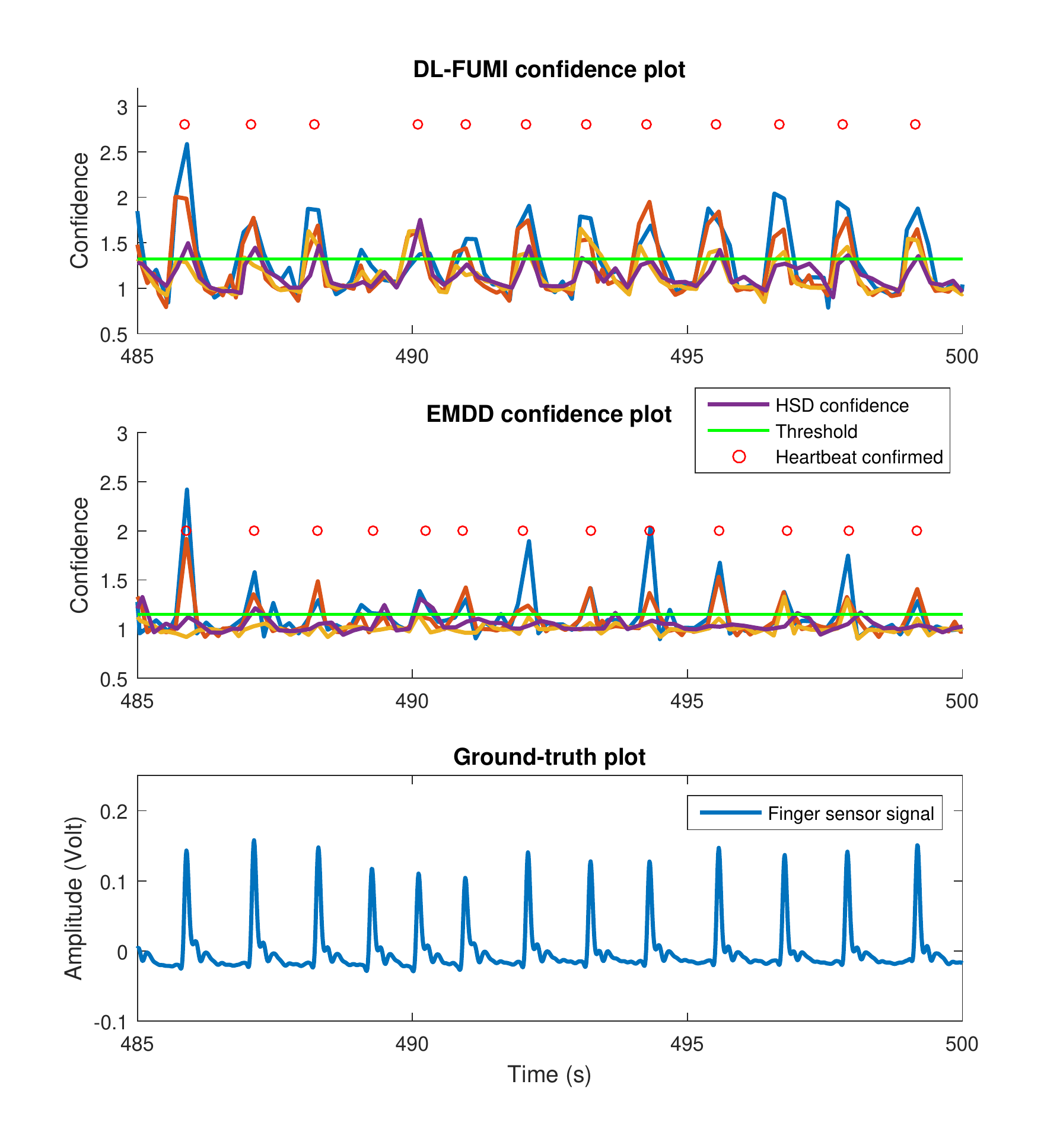}
		\caption{Confidence Value and Confirmed Heartbeat. The confidence estimated by DL-FUMI (upper) is more consistent than that estimated by EMDD (middle), thus provides more robust detection performance.}\label{fig:detc_confid}
	\end{center}
\end{figure}

For each subject, 10 minutes of their BCG signal was split into 5 minutes for training and 5 minutes for testing. For DL-FUMI, the parameters used for all the subjects are $T=3$, $M=3$, $\lambda=5\times10^{-3}$, $\Gamma=5\times10^{-3}$ and $\beta=90$. Fig. \ref{fig:HB_concepts_DLFUMI} shows estimated heartbeat concepts for subject No. 10 as an example, where we can see the heartbeat concept estimated by DL-FUMI maintains prominent J-peaks. For comparison, we applied EMDD \cite{Zhang:2002} which is a widely used multiple instance concept learning algorithm to the same data and show the estimated heartbeat concept in Fig. \ref{fig:HB_concepts_emdd}. From Fig. \ref{fig:HB_concepts_DLFUMI} and \ref{fig:HB_concepts_emdd} we can clearly see that although both the heartbeat concepts estimated by DL-FUMI and EMDD have prominent J-peaks, DL-FUMI is able to learn multiple concepts to account for the variability in heartbeat prototype during sleeping, which helps improve performance in heartbeat detection and rate estimation. {In order to illustrate the advantages of proposed DL-FUMI over our previously proposed $e$FUMI algorithm \cite{jiao2016heart, jiao2016multipleicpr}, the heartbeat estimated by $e$FUMI from the same subject is shown in Fig. \ref{fig:HB_concepts_eFUMI}. From Fig. \ref{fig:HB_concepts_eFUMI} we can clearly see that although the heartbeat concept estimated by $e$FUMI preserves a prominent J-peak, it does not have the fidelity of the heartbeat signals that DL-FUMI provides.  This is because $e$FUMI assumes each signal is a convex combination of target and/or non-target concepts (as it was originally developed for hyperspectral image analysis), which may not be suitable for the heartbeat signals. The estimated heartbeat concept from this convex model will be the vertex of the assumed convex hall which falls outside the data and does not provide a true estimation of the subject's heartbeat concept. Furthermore, the $e$FUMI only estimates a single target concept which cannot account for the large variability in the target class.   As a comparison, the multiple heartbeat concepts estimated by DL-FUMI shown in Fig. \ref{fig:HB_concepts_DLFUMI} provide better representations of the heartbeat signal. }
\begin{figure}
	\vspace{-5mm}
	\begin{center}
		\includegraphics[width=8.5cm]{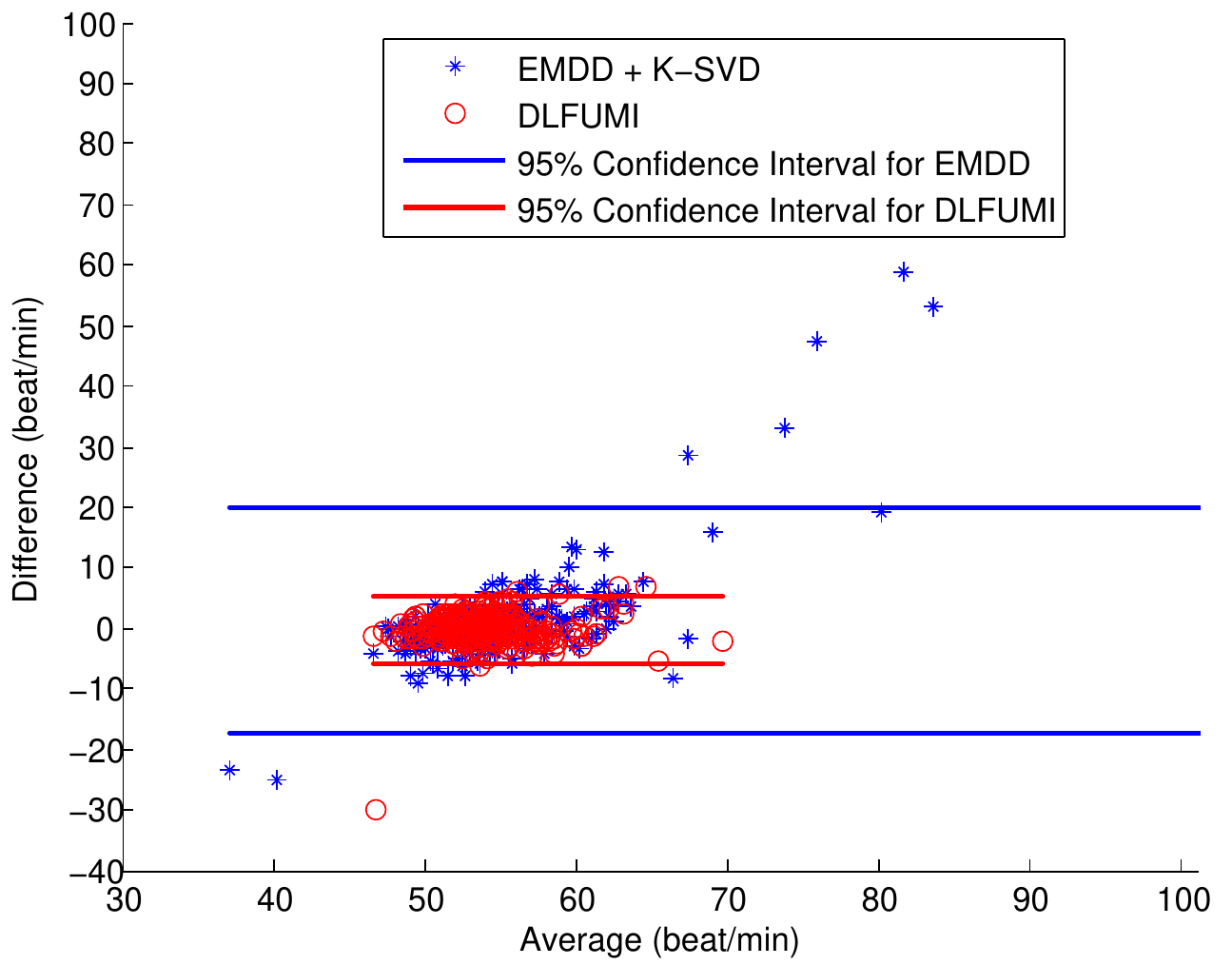}
		\caption{The Bland Altman plot comparison of DL-FUMI and EMDD for Subject No. 10. The outlier by DL-FUMI (red circle in the bottom left)  corresponding to the missed detection appeared around 489s shown in Fig. \ref{fig:detc_confid}, but the plot of DL-FUMI is overall more compact and has less false alarms as well as missed detections than EMDD.}\label{fig:BA_plot}
	\end{center}
\end{figure}

\begin{figure}
	\begin{center}
		\subfigure[DL-FUMI]{
			\includegraphics[width=8cm]{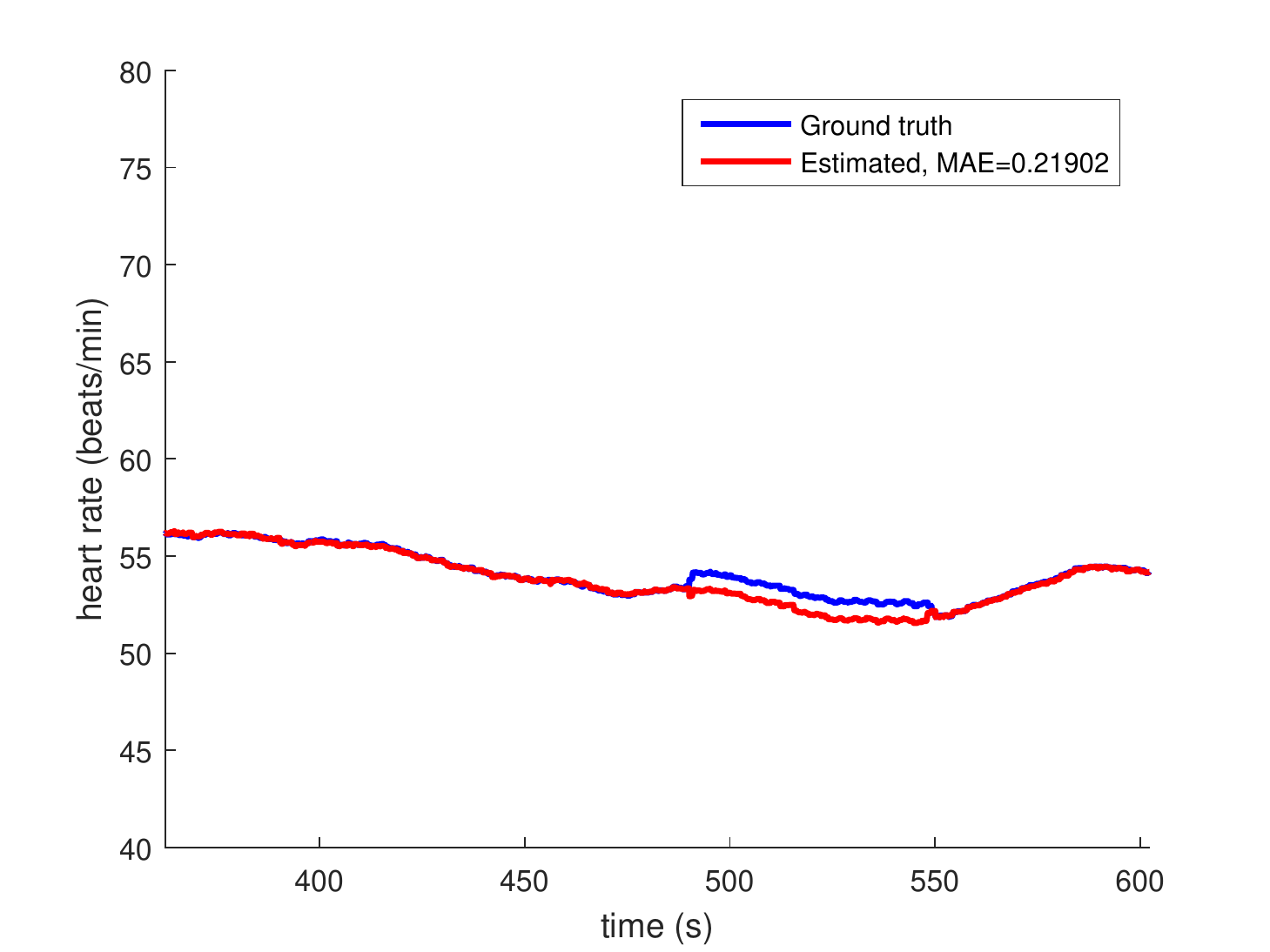} \label{fig:rate_DLFUMI}}
		\subfigure[EMDD]{
			\includegraphics[width=8cm]{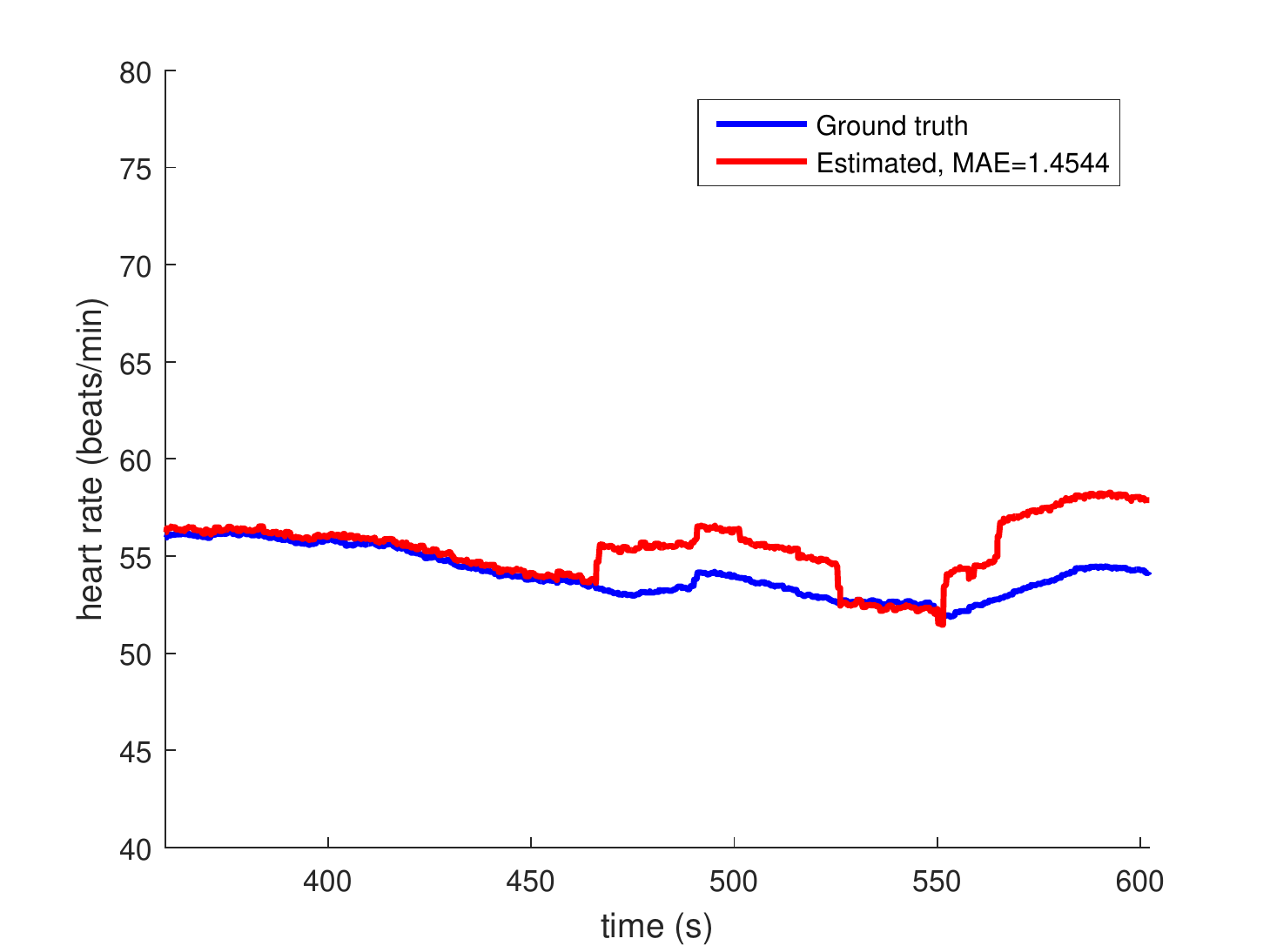} \label{fig:rate_EMDD}}
		\caption{Heart rate estimation using 1 minute sliding window. (a) Heart rate estimated by DL-FUMI, where the drop around 500s resulted from one missed detection at 489s. (b) Heart rate estimated by EMDD, several false alarms account for the raise in heart rate estimation. }\label{fig:HB_rate}
	\end{center}
\end{figure}

After learning heartbeat concepts, heartbeat detection on test data can be carried out. In the results shown in this paper, the HSD was applied to the test data to get a confidence value for each data point to be a true heartbeat signal. Since EMDD only learns a single target concept, we applied K-SVD \cite{aharon2006rm} which is a widely used unsupervised dictionary learning algorithm to the negative labeled training bags to get a set of non-target concepts for EMDD heartbeat detection using HSD.  Fig. \ref{fig:detc_confid} shows the excerpt of HSD 4 - channel confidence value (485s - 500s) for subject No. 10  estimated by the heartbeat concepts of DL-FUMI (shown in Fig. \ref{fig:HB_concepts_DLFUMI}) and EMDD (shown in Fig. \ref{fig:HB_concepts_emdd}), respectively.  In this procedure, a {heartbeat (J-peak) is confirmed through a voting procedure requiring at least two confidence values within a neighborhood (25 samples) that are greater than a threshold (1.32) across all four transducers.} The neighborhood and threshold values are determined via cross-validation on training data.  From Fig. \ref{fig:detc_confid} we can see that the confidence {peaks} estimated by DL-FUMI match the peaks of the finger {sensor} signal very well. A {missed} detection was found at about 489s. For comparison, the confidence {estimated by EMDD is not as} prominent as DL-FUMI and there are several false alarms found. Fig. \ref{fig:BA_plot} shows the beat to beat Bland Altman plot of DL-FUMI and EMDD, where we can see the plot of DL-FUMI is more compact than that of EMDD. For DL-FUMI, there is one outlier in the bottom left corresponding to the {missed} detection appeared around 489s in Fig. \ref{fig:detc_confid}. {The $95\%$ confidence interval for the Bland Altman plot is also shown in Fig. \ref{fig:BA_plot}, where the $95\%$ confidence interval achieved by DL-FUMI is much narrower than that of EMDD. }

\begin{table*}
		\setlength{\tabcolsep}{5pt}
	\begin{center}	
		\caption{Performance of DL-FUMI and Comparisons Across the 40 Subjects, bold for the best, underline for the second best, standard deviations smaller than 0.01 are denoted as 0.00}\label{tab:heart_rate_error}
		\begin{tabular}{lllllllll}
			\Xhline{1.2pt}
			\multirow{2}{*}{Subject} &  {  Age, sex (F/M), BMI} & \multicolumn{5}{c} {MAE (beat/min)}  \\
			\Xcline{3-9}{0.5pt}
			&  {weight (kg), height (cm) }& DL-FUMI & EMDD$+$K-SVD & EN & WPPD & CA & HT & $e$FUMI\\
			\Xhline{0.7pt}
			1 & 43, M, 23.3, 74.8, 179  &    \textbf{0.19}$\pm$0.00     &  0.96$\pm$0.82       &	 \underline{0.22} &    1.26   &1.76$\pm$0.09        &	 {0.79}  & 0.28$\pm$0.02\\
			2 & 21, M, 23.1, 74, 179    &    {0.85}$\pm$0.05   &  4.99$\pm$1.48       &  \textbf{0.58} &    1.67   &1.68$\pm$0.14      &	 {0.94}  &\underline{0.64}$\pm$0.15\\
			3 & 33, M, 27.2, 75, 166    &    \underline{0.22}$\pm$0.00    &  0.71$\pm$0.94       &  \textbf{0.17} &  1.02   &1.05$\pm$0.03     &    {0.77}   & \textbf{0.17}$\pm$0.09 \\		
			4 & 23, F, 21.1, 52, 157    &    \underline{0.96}$\pm$0.57  &  3.55$\pm$1.76       &	 14.69 &    1.28   &1.24$\pm$0.03      &	 \textbf{0.85}   & 1.10$\pm$0.24\\		
			5 & 28, M, 37.9, 127, 183   &    \textbf{0.83}$\pm$0.33      &  4.83$\pm$0.89       &  2.84   &    1.25   &1.10$\pm$0.07    &	 {1.00}   &  \underline{0.84}$\pm$0.20\\		
			6 & 21, M,  22.2, 75, 184   &   \textbf{0.45}$\pm$0.07       &  2.61$\pm$1.61       &	 20.20   &    1.26   &1.10$\pm$0.03   &	 {0.66}   &  \underline{0.59}$\pm$0.38\\		
			7 & 32, M, 30.1, 82, 165    &    {1.28}$\pm$0.19   &  4.45$\pm$2.23       &  1.83   &    1.47   &1.98$\pm$0.09   &	 \textbf{0.79}   &  \underline{1.21}$\pm$0.05\\		
			8 & 32, M, 27.2, 92, 180    &    {1.46}$\pm$0.89          &  3.47$\pm$2.72       &  1.65    &  \underline{1.04}   &1.06$\pm$0.04  & \textbf{0.48}  &  1.40$\pm$0.75\\		
			9 & 29, M, 19.7, 68, 186    &     2.47$\pm$0.60           &  1.34$\pm$0.46       &	 \textbf{0.77}    &    2.12   &2.03$\pm$0.04    &	 \underline{1.26}  &  1.79$\pm$0.27\\		
			10 & 24, F, 22.8, 62, 165   &    \textbf{0.24}$\pm$0.00      &  3.00$\pm$1.72       &	\underline{0.53}     &    1.07   &1.39$\pm$0.03  &	 {0.64}  &  4.14$\pm$0.59\\		
			11 & 31, M, 22.5, 68, 174   &    {1.09}$\pm$0.00           &  2.16$\pm$1.20       &	 \textbf{0.61}   &    1.66   &1.42$\pm$0.04    &	 \underline{1.05}   &  8.66$\pm$1.35\\
			12 & 27, M, 24.2, 70, 170   &    \underline{0.30}$\pm$0.10    &  0.66$\pm$0.55       &	 \textbf{0.18}    &    1.28   &1.10$\pm$0.04   &	 {0.94}   &  6.34$\pm$5.48\\
			13 & 29, M, 23.6, 79, 183   &    {0.27}$\pm$0.37     &  1.14$\pm$0.28       &	 \underline{0.09}    &    1.03   &1.12$\pm$0.04    &	 {0.53}   &  \textbf{0.07}$\pm$0.04\\		
			14 & 18, M, 25.6, 83, 180   &    {1.64}$\pm$0.37     &  2.88$\pm$1.52       &	 6.52    &    {1.63}   &2.27$\pm$0.04   &	 \underline{1.62}  &  \textbf{1.26}$\pm$0.29\\		
			15 & 39, M, 25.0, 74, 172   &    \textbf{0.14}$\pm$0.00      &  0.45$\pm$0.33       &	 {0.27}    &    1.45   &1.29$\pm$0.04   &	 {0.89}  &  \underline{0.20}$\pm$0.00\\		
			16 & 26, M, 21.2, 65, 175   &    \textbf{0.22}$\pm$0.02     &  2.09$\pm$1.14       &	 6.12   &    1.25   &1.65$\pm$0.05    &	 \underline{0.50}  &  1.97$\pm$0.71\\		
			17 & 31, M, 28.9, 100, 186  &   \underline{0.10}$\pm$0.00      &  1.66$\pm$1.12       &	 0.49    &    1.04   &1.87$\pm$0.19    &	 {0.46}   &  \textbf{0.08}$\pm$0.00\\		
			18 & 27, M, 22.3, 70, 177   &    \underline{0.45}$\pm$0.01  &  1.64$\pm$1.72       &	\textbf{0.20}   &    1.21   &1.07$\pm$0.04   &	 {0.63}  &  10.28$\pm$4.52\\		
			19 & 30, M, 25.1, 76, 174   &    \textbf{0.15}$\pm$0.00    &  1.18$\pm$0.95       &	 {0.62}   &    1.45   &1.48$\pm$0.05   &	 {0.71}  &  \underline{0.16}$\pm$0.00\\		
			20 & 23, M, 24.7, 73, 172   &    \textbf{0.56}$\pm$0.03     &  4.43$\pm$3.93       &	 \underline{0.85}    &    2.35   &2.12$\pm$0.09   &	 {1.42}  &  1.25$\pm$0.53\\	
			21 & 30, F, 19.7, 57, 170   &    {1.18}$\pm$0.37         &  4.08$\pm$1.55       &	 4.29    &    \underline{0.90}   &1.07$\pm$0.14   &	 \textbf{0.23}  &  3.30$\pm$0.51\\
			22 & 27, M, 27.5, 86, 177   &    \textbf{0.10}$\pm$0.00     &  2.12$\pm$0.34       &	\underline{0.38}   &    1.65   &1.51$\pm$0.07   &	 {0.44}  &  2.49$\pm$0.80\\
			23 & 24, F, 20.8, 60, 170   &    \underline{0.81}$\pm$0.00    &  1.76$\pm$0.72       &	 7.65    &    1.44   &1.61$\pm$0.13   &	 {1.12} &  \textbf{0.73}$\pm$0.21 \\		
			24 & 25, M, 27.1, 86, 178   &    \textbf{0.32}$\pm$0.03    &  2.21$\pm$0.27       &	 7.21    &    2.74   &1.86$\pm$0.08   &	 {1.14}   &  \underline{0.50}$\pm$0.12\\		
			25 & 49, M, 23.0, 83, 190   &    {0.20}$\pm$0.04    &  0.73$\pm$0.75       &	 \underline{0.07}    &    1.35   &1.41$\pm$0.04    &	 {0.80}  &  \textbf{0.04}$\pm$0.01\\		
			26 & 22, M, 26.0, 92, 188   &   \textbf{0.25}$\pm$0.02     &  {2.08}$\pm$1.82     &	 \underline{0.49}    &    2.13   &2.30$\pm$0.26    &    1.81   &  0.59$\pm$0.07\\		
			27 & 33, M, 29.2, 82, 167.6 &  \textbf{0.56}$\pm$0.27    &  4.57$\pm$3.26       &	 {0.89}    &    1.45   &1.49$\pm$0.04    &	 {1.00} &  \underline{0.80}$\pm$0.39 \\		
			28 & 28, M, 21.2, 73, 185.4 &  \textbf{0.62}$\pm$0.22     &  1.64$\pm$1.10       &	 3.21    &    1.72   &1.96$\pm$0.08   &	 \underline{1.38}   &  5.36$\pm$1.00\\		
			29 & 34, M, 25.6, 84, 181   &    0.47$\pm$0.08        &  1.15$\pm$0.54       &	 \textbf{0.09}   &    0.89   &0.87$\pm$0.06  &	 \underline{0.34}   &  1.61$\pm$1.04\\		
			30 & 34, M, 24.4, 64, 162   &    \textbf{0.15}$\pm$0.03    &  3.67$\pm$3.14       &  \underline{0.17}    &    1.06   &1.28$\pm$0.16   &	 {0.34}  &  0.24$\pm$0.08\\	
			31 & 32, F, 28.2, 77, 165.1 &  \textbf{0.99}$\pm$0.12     &  3.75$\pm$1.31       &  1.47   &     \underline{1.30}   &2.36$\pm$0.13   &	 {1.71}  &  2.36$\pm$2.23\\
			32 & 22, F, 21.0, 51, 156   &    \underline{1.26}$\pm$0.57     &  2.53$\pm$1.38       &	 10.23   &     1.59   &2.36$\pm$0.04   &	 \textbf{1.03}   &  \textbf{1.03}$\pm$1.39\\
			33 & 27, M, 23.8, 77, 180   &     \underline{0.71}$\pm$0.07    &  0.73$\pm$0.23       &  8.43     &    1.82   &0.98$\pm$0.01  &	 \textbf{0.54}   &  0.80$\pm$0.20\\		
			34 & 26, M, 23.7, 83, 187   &    \textbf{0.61}$\pm$0.13    &  2.72$\pm$2.27       &	 4.65    &    1.49   &1.55$\pm$0.08   &	 \underline{0.64}  &  7.27$\pm$4.50\\		
			35 & 28, F, 18.3, 48, 162   &    \textbf{0.15}$\pm$0.05      &  3.98$\pm$1.63       &	\underline{0.43}   &    1.34   &1.15$\pm$0.03   &	 {0.76}  &  1.13$\pm$0.03\\		
			36 & 38, M, 36.9, 120, 180.3 &    {1.44}$\pm$0.31 	  &  {1.00}$\pm$0.52     &	 1.68    &    1.15   &1.19$\pm$0.16   &	 \underline{0.47}  &  \textbf{0.35}$\pm$0.02\\		
			37 & 37, M, 22.1, 68, 175.3 &    \textbf{0.23}$\pm$0.00    &  2.42$\pm$1.54       &	 0.85     &    0.74   &0.89$\pm$0.05    &	 \underline{0.40}  &  0.48$\pm$0.41\\		
			38 & 29, M, 25.8, 79, 175   &    \textbf{0.46}$\pm$0.16  	 &  1.37$\pm$1.24       &	 6.53      &    1.13   &0.87$\pm$0.07   &	 \underline{0.47}  &  0.51$\pm$0.09\\		
			39 & 23, M, 23.0, 68, 172   &    {0.71}$\pm$0.38	   &  4.11$\pm$3.43       &	 \underline{0.52}      &    1.80   &1.84$\pm$0.08    &	 {1.00}  &  \textbf{0.51}$\pm$0.33\\		
			40 & 32, M, 30.9, 99, 179   &    \textbf{0.37}$\pm$0.04  	  &  2.57$\pm$2.10       &  1.94    &    1.24   &1.44$\pm$0.08   &	 \underline{0.55}  &  0.94$\pm$0.10\\	
			\Xhline{0.7pt}
			\multicolumn{2}{c} {Total average}  & \textbf{0.64}                  & 2.43              &  3.02    &     1.42   &  1.49           & \underline{0.83}   &  1.84\\
			\Xhline{1.2pt}
		\end{tabular}
	\end{center}
\end{table*}

\begin{table}
	\setlength{\tabcolsep}{5pt}
	\begin{scriptsize}
	\caption{The Correlation Coefficients between Performance and Age, Weight, Height, BMI and Groundtruth.} \centering
	\begin{tabular}
		{ c c c c c c c c }\hline
		& DL-FUMI & EMDD & WPPD & CA & EN & HT &$e$FUMI \\\hline
		Age & -0.20 & -0.37 & -0.28 & -0.31 & -0.46 & -0.35 & -0.17\\
		Weight & 0.04 & -0.06 & -0.12 & -0.04 & -0.21 & -0.01 & -0.19\\
		Height & 0.00 & -0.35 & -0.07 & -0.01 & -0.11 & 0.01 & 0.03\\
		BMI & 0.05 & 0.10 & -0.08 & -0.04 & -0.21 & -0.01 &-0.24\\
		average GT  &  0.24    &  -0.01    & 0.03     &  -0.05     &   0.46    & -0.11 &  -0.10    \\\hline
	\end{tabular}
	\label{table:performance comparison of age, weight, height, and BMI}
	\vspace{-5mm}
	\end{scriptsize}
\end{table}

{For heart rate estimation, the average of beat-to-beat heart rates over 1 minute is computed using a sliding window. The mean absolute error (MAE) in heart rate is computed from the difference between the estimation and the beat-to-beat basis of the finger sensor for all subjects.} Fig. \ref{fig:rate_DLFUMI} and Fig. \ref{fig:rate_EMDD} show the estimated heart rate and errors on testing data for subject No. 10 by DL-FUMI and EMDD, respectively. From Fig. \ref{fig:rate_DLFUMI} we can see that the drop in estimated heart rate around 489s comes from the {missed} detection of heartbeat in Fig \ref{fig:detc_confid}. As a comparison, the rise in estimated heart rate in Fig. \ref{fig:rate_EMDD} comes from the several false alarms estimated by EMDD.

{A comprehensive study conducted on 40 subjects and comparison with EMDD, $e$FUMI and state-of-the-art BCG heart rate monitoring algorithms, window-peak-to-peak (WPPD) \cite{Heise2010}, clustering approach (CA) \cite{rosales2012heartbeat}, energy (EN) \cite{Lydon2015}, and Hilbert transform (HT) \cite{Su:2012,Su:2017} are presented. {For a fair comparison with other methods that are not beat-to-beat based, for example, the HT algorithm uses frequency domain processing and gives an average heart rate estimate over a one minute segment  for every 15s \cite{Su:2017}, the one minute sliding window was advanced forward at every 15s when generating the results for comparison. The detail of the performance comparison is listed in Table I. The error shown in Table I is the MAE between the estimation and groundtruthing at every 15s interval,} where the bold numbers highlight the best and underlined numbers indicate the second best performance for each subject. Over the 40 subjects, the MAE of proposed algorithm is 0.64 (beat/min), which is the best over the comparison algorithms. The overall MAE of EMDD+K-SVD, EN, WPPD, CA, HT and $e$FUMI are 2.43, 3.02, 1.42, 1.49, 0.83 and 1.84 (beat/min), respectively.} For EN, there are 3 subjects that have aberrant estimation (with error greater than 10) which make the average result worse. If we ignore these three outliers, the overall MAE of EN is 2.04 (beat/min). 

\begin{figure}
	\begin{center}
		\includegraphics[width=9.3cm]{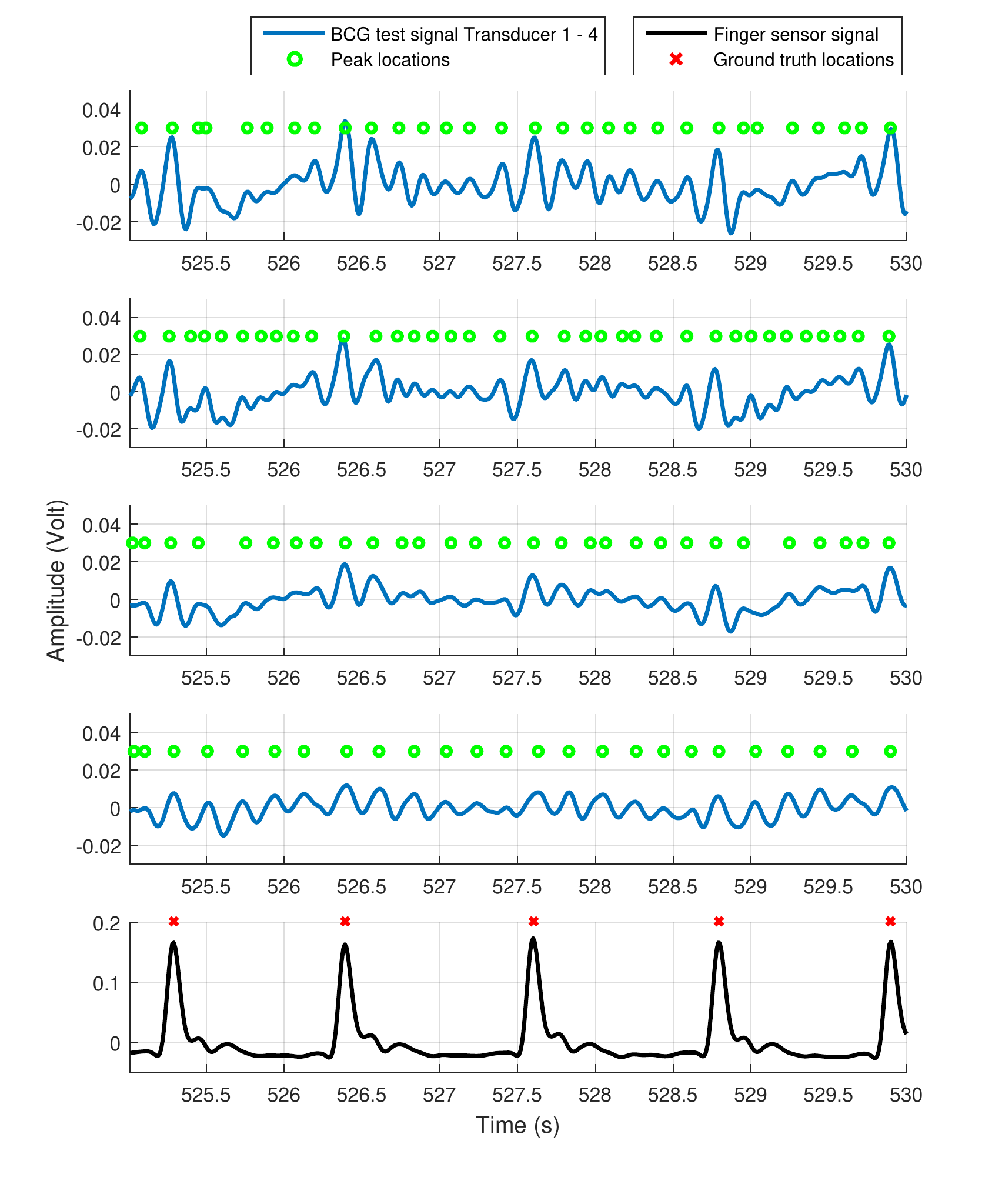}
		\caption{The 5s heartbeat signal plot of subject No. 10 with relatively large variability in heart rate.}\label{fig:5s_sig_plot}
	\end{center}
	\vspace{-3mm}
\end{figure} 

To examine if the performance of proposed algorithm is related to subject's age, weight, height, BMI, and magnitude of subject's heart rate, we compute the correlation coefficient between the results in Table~\ref{tab:heart_rate_error} and subjects' age, weight, height, BMI and the average of groundtruth heart rate over testing data (denoted as average 
GT). The sample Pearson correlation coefficient $r$ defined in Eq. \eqref{eqn:corr_coef} measures the linear correlation between two variables $x$ and $y$, where $i$ is the index of the variables and $\bar{x}$ and $\bar{y}$ are the mean of two variables, respectively. Table~\ref{table:performance comparison of age, weight, height, and BMI} shows the corresponding correlation coefficient of the proposed algorithm and the other five comparison algorithms. For weight, height, BMI and heart rate magnitude, all seven algorithms show no significant relation. For age, all algorithms consistently show negative relationship with error, this is mainly due to the increase in heartbeat variations with the decrease in age. However, compared with the other five algorithms, the proposed DL-FUMI is more robust to age since DL-FUMI is able to learn multiple concepts to account for heartbeat variability. Based on the results, we can tell that the performance is not related to age, weight, height, and BMI, which validates the robustness of the proposed algorithm.

\begin{equation}
r=\frac{\sum_{i}(x_i-\bar{x})(y_i-\bar{y})}{\sqrt{\sum_{i}(x_i-\bar{x})^2}\sqrt{\sum_{i}(y_i-\bar{y})^2}},
\label{eqn:corr_coef}
\end{equation}

\begin{figure}
	\begin{center}
		\subfigure[DL-FUMI Beat-to-beat Detection]{
			\includegraphics[width=8cm]{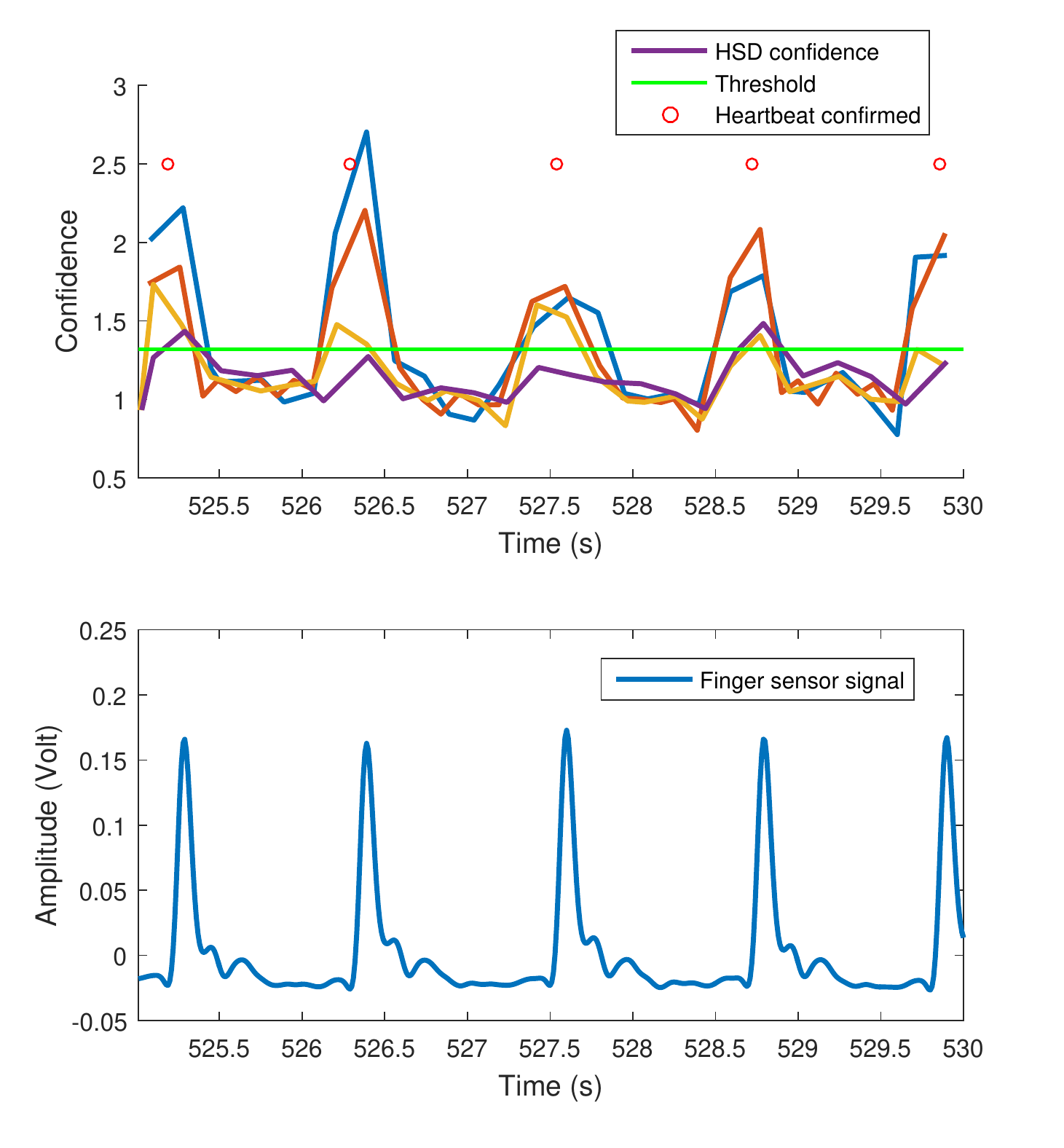} \label{fig:det_5s_DLFUMI}}
		\subfigure[Heart Rate Estimates]{
			\includegraphics[width=8cm]{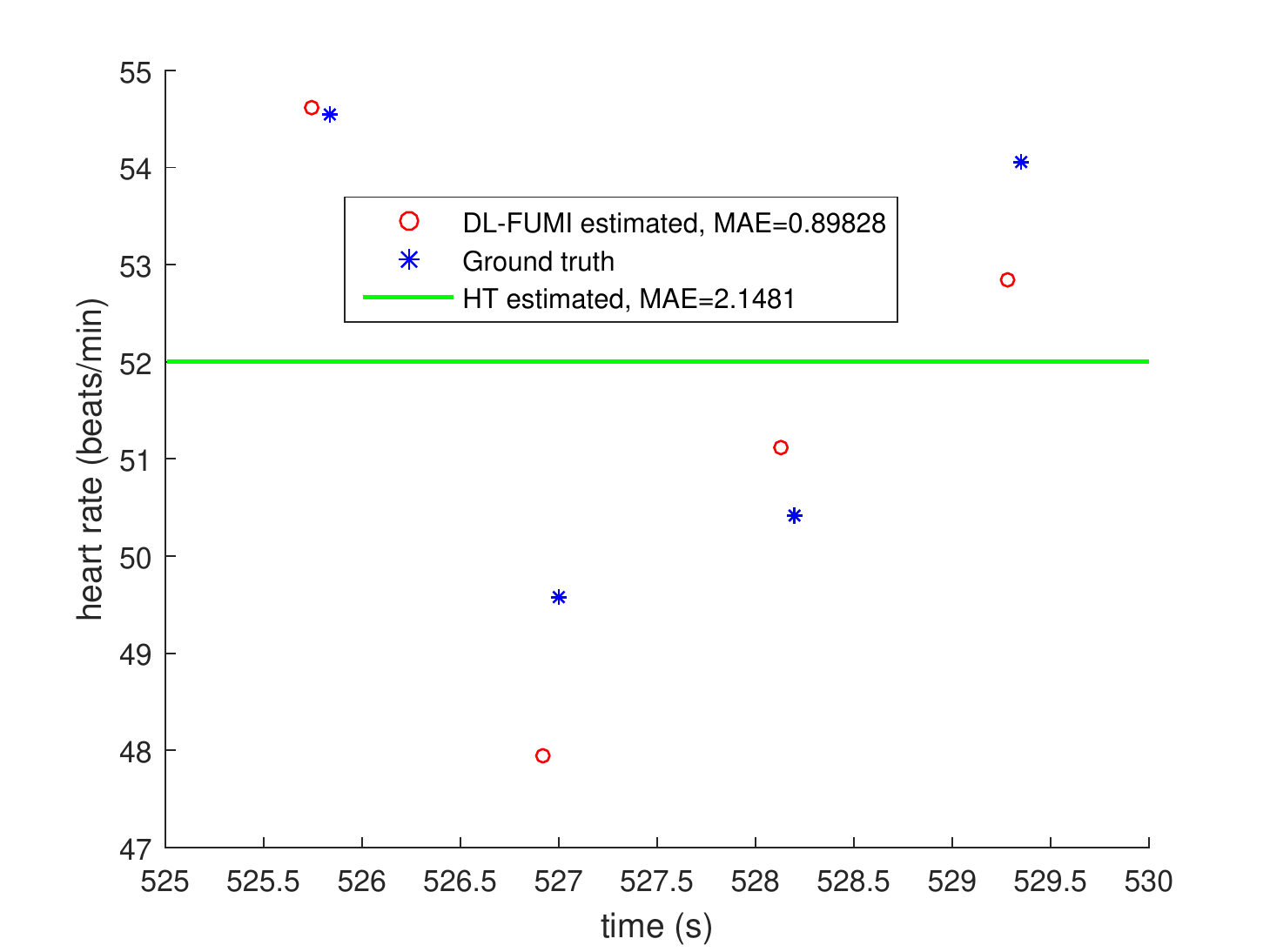} \label{fig:5s_rate_DLFUMI_HT}}
		\caption{Heart Rate Estimation from 5s signal. (a)  Beat-to-beat Detection by DL-FUMI (upper) and Finer Sensor Plot (ground truth, middle). (b) Beat-to-beat Heart Rate estimated by DL-FUMI (red circles) and finger sensor (blue stars), where the comparison HT algorithm only estimates a constant heart rate (green line) from a period of signal.}\label{fig:det_5s}
	\end{center}
\end{figure}

{The HT algorithm obtains an overall performance close to the proposed algorithm (0.19 worse than DL-FUMI in MAE). To further evaluate the difference between the two statistics, a T-test for paired data \cite{student1908probable, kutner2005applied} was applied to compare the performance of DL-FUMI and HT shown in Tab. \ref{tab:heart_rate_error}.  The T-statistics between DL-FUMI and HT is -2.16. Compared with the difference in the sample average which is -0.19, DL-FUMI performs much better than HT under this metric. Also the hypothesis ``The two means are the same'' ($\mathbf{H}_0$) is rejected at $5\%$ significance level, which indicates the advantage of DL-FUMI over HT.}



{Furthermore, HT computes the heart rate from the frequency domain and only estimates an overall heart rate from a segment of signal, thus does not capture each heartbeat presence as well as the heart rate variability over time.  In comparison, the proposed algorithm estimates a set of personalized heartbeat concept that provides a concrete  characterization of one's heartbeat pattern. After learning the heartbeat concept, a real time heartbeat detection and beat-to-beat heart rate estimation can be conducted. To further evaluate the two algorithms, we select just 5s signal segment with relatively large heart rate variability from subject No. 10 shown in Fig. \ref{fig:5s_sig_plot}. The beat-to-beat detection by DL-FUMI is shown in Fig. \ref{fig:det_5s_DLFUMI} and the heart rates estimated by both DL-FUMI and HT are shown in Fig \ref{fig:5s_rate_DLFUMI_HT}. From Fig. \ref{fig:5s_rate_DLFUMI_HT} we can see that if the subject has high heart rate variation, the HT cannot follow the variation well. In contrast, the beat-to-beat heartbeats detected by DL-FUMI provide a real time tracking of the heart rate variability within a short period of time. }

\subsection{Heartbeat Detection and Rate Estimation from Batch Training}\label{sec:exp_batch_training}
\begin{figure}
	\begin{center}
		\includegraphics[width=9cm]{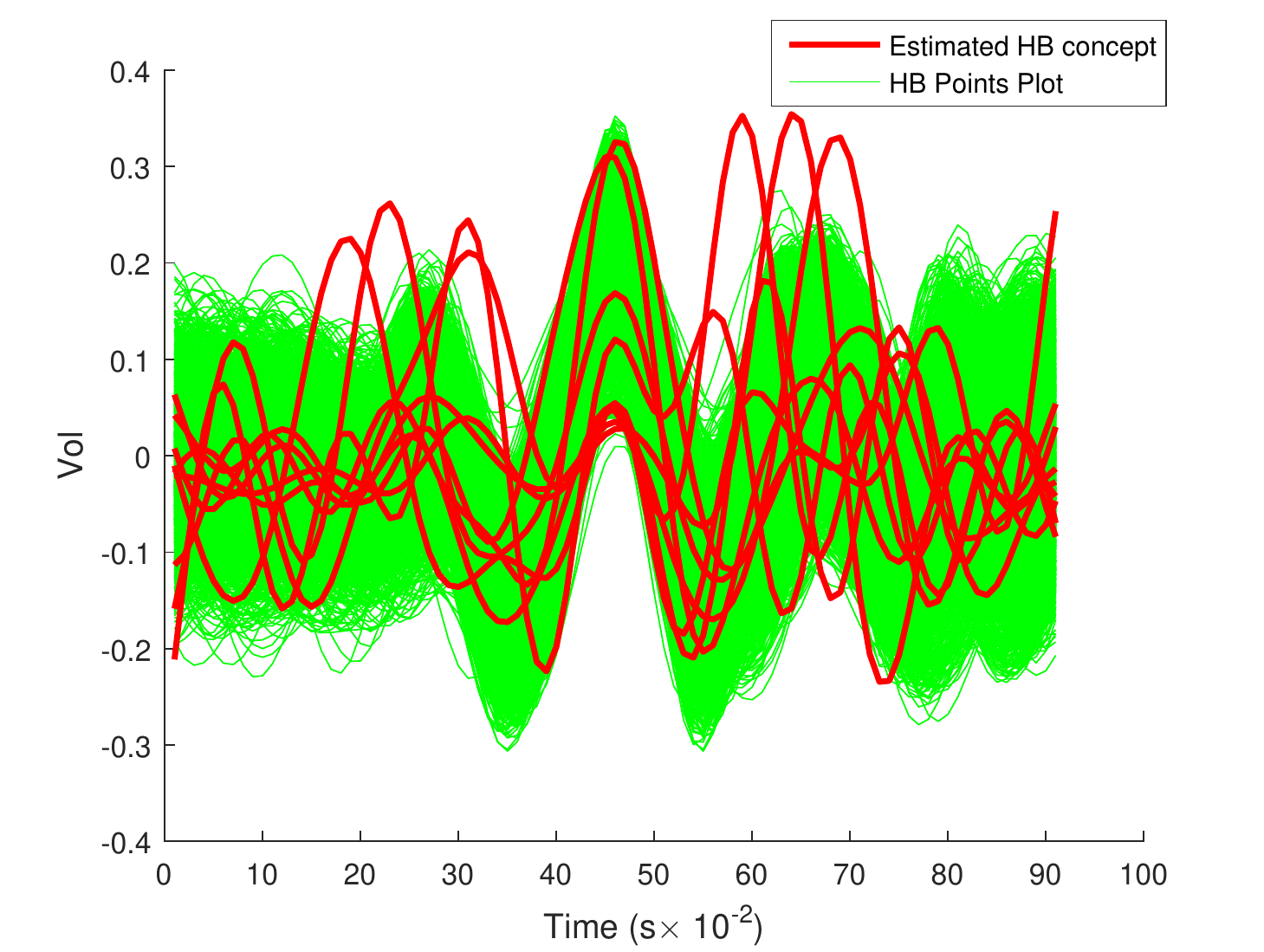}
		\caption{ Heartbeat concept estimated by DL-FUMI from batch training on second twenty subjects. }\label{fig:DLFUMI_concept_batch}
	\end{center}
\end{figure}

\begin{figure}
	\begin{center}
		\includegraphics[width=9cm]{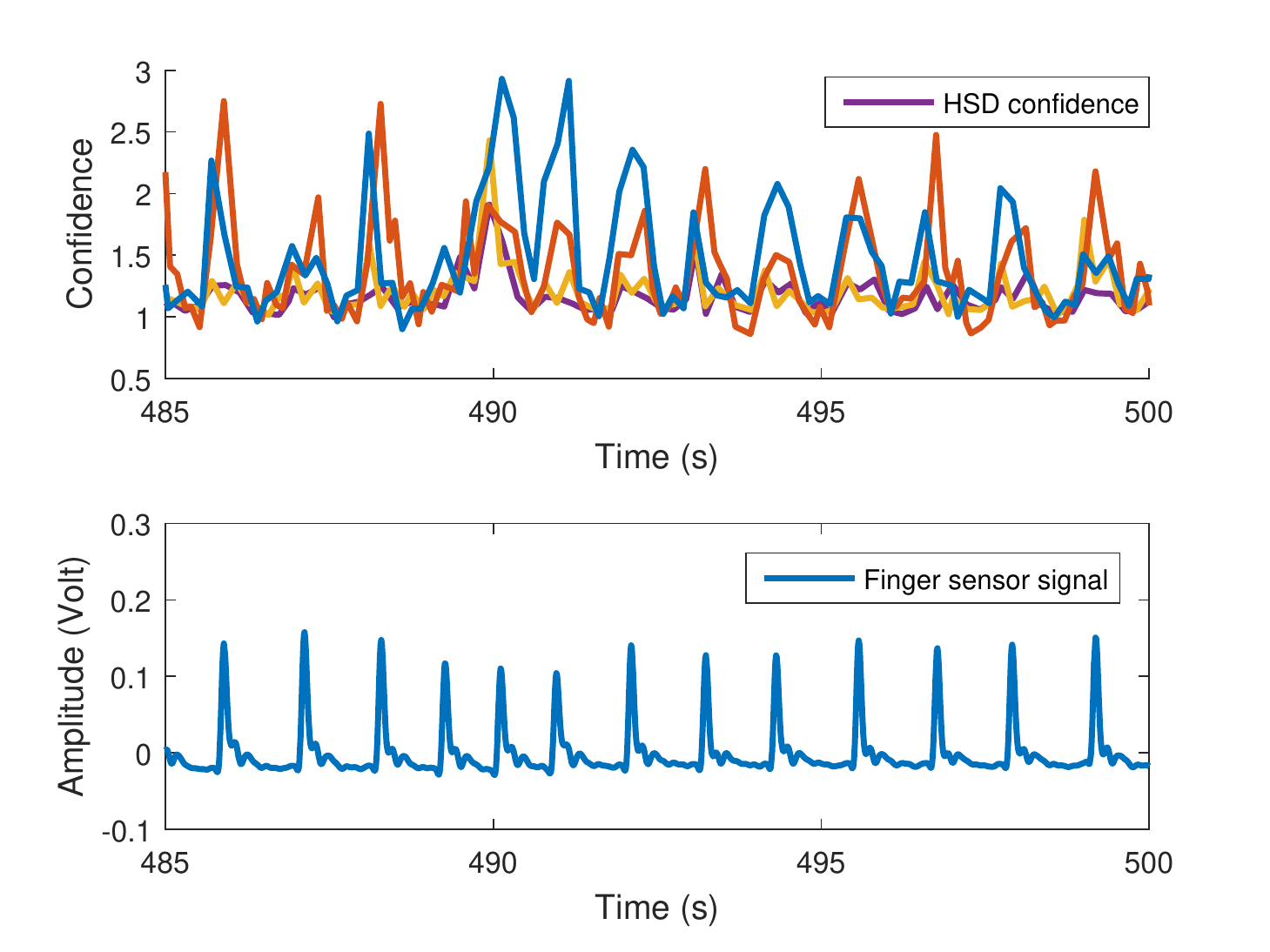}
		\caption{  Confidence value estimated for subject No. 10 using the heartbeat concept shown in Fig. \ref {fig:DLFUMI_concept_batch}}\label{fig:confid_batch_training}
	\end{center}
\end{figure}

{In order to show heartbeat detection performance of DL-FUMI without requiring training for each subject individually, the 40 subjects were split into two groups and cross validated for training and testing. Specifically, the five minutes of training BCG signal from the first twenty subjects were grouped collectively as training data and tested individually on the respective five minutes of testing BCG signal of the second twenty subjects; then vice versa. The DL-FUMI parameters used for the batch training are $T=9$, $M=9$, $\lambda=1\times10^{-3}$, $\Gamma=5\times10^{-3}$ and $\beta=120$. Here, the assumed number of target and non-target concepts were increased to 9 to account for the large variability from the batch training data and train a more generally representative heartbeat concept set. The estimated heartbeat concepts from batch training on the second twenty subjects are shown in Fig. \ref{fig:DLFUMI_concept_batch}. Fig. \ref{fig:confid_batch_training} shows the confidence value estimated for subject No. 10 using the hybrid detector with heartbeat concepts estimated from the batch training on the second 20 subjects. }
	
{For estimation of the heart rate without learning the thresholding parameters individually, frequency domain analysis was performed on the estimated confidence value. Specifically, the discrete Fourier transform was applied to the 1 minute long confidence value at every 15s interval. Then the frequency corresponding to the biggest component magnitude of the 1 minute confidence segments from the 4 transducers was adopted as the heart rate frequency, and transformed into beat/min as the estimated heart rate for this 1 minute segment. The overall MAE estimated using this batch training and frequency analysis for the entire forty subjects is 0.80, which is slightly worse than learning heartbeat concept and estimating heart rate individually, but is still better than the performance of HT algorithm and other comparison approaches.  }

\begin{figure}
	\begin{center}
		\includegraphics[width=9.3cm]{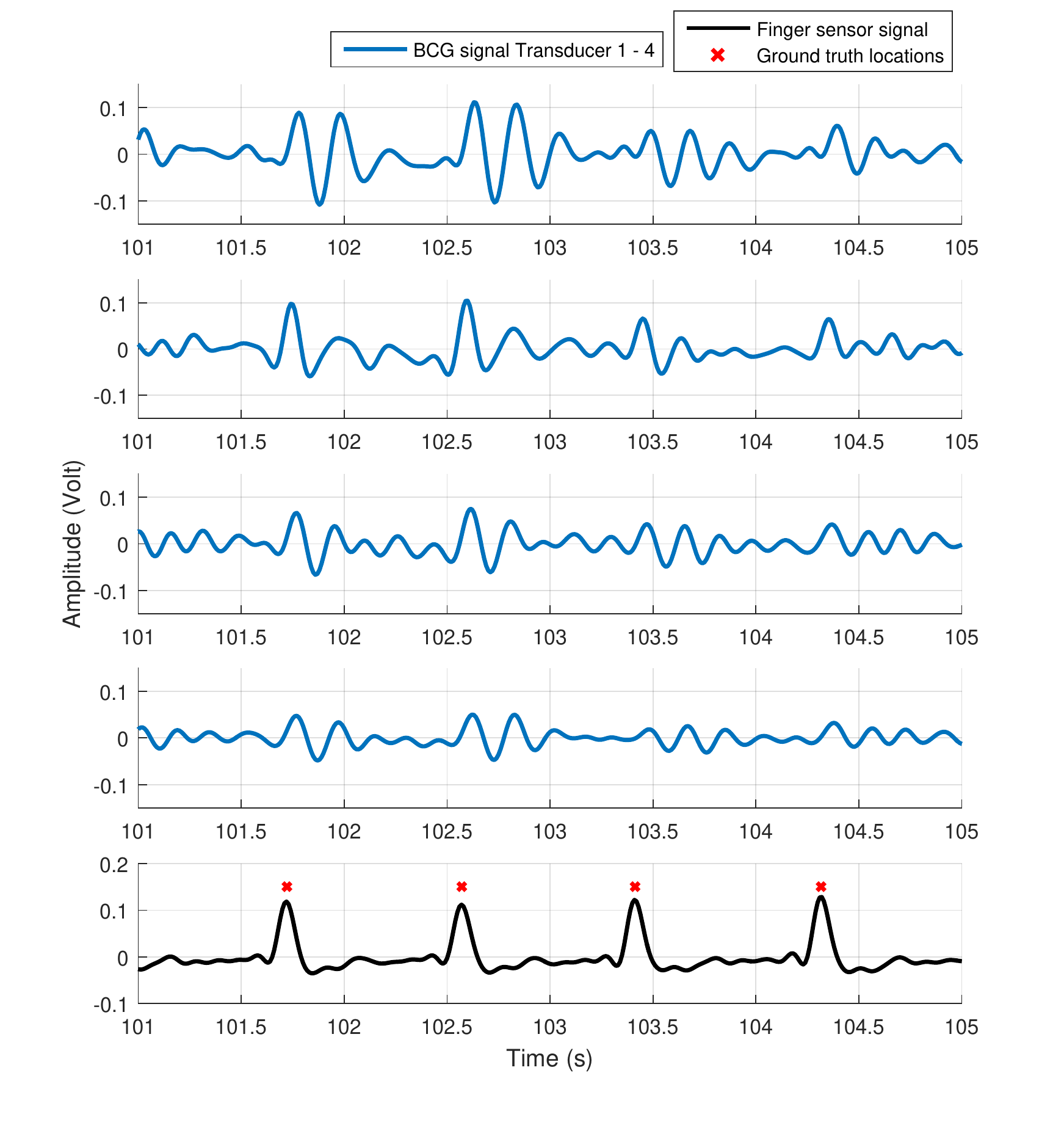}
		\caption{ Plot of BCG exercise signal and ground truth for subject No. 10. Compared with the resting data shown in Fig. \ref{fig:BCG_FS_plot}, exercise data maintains much larger signal magnitude and shorter J-J perk interval.}\label{fig:BCG_sig_exercise}
	\end{center}
\end{figure}

\begin{figure}
	\begin{center}
		\includegraphics[width=8cm]{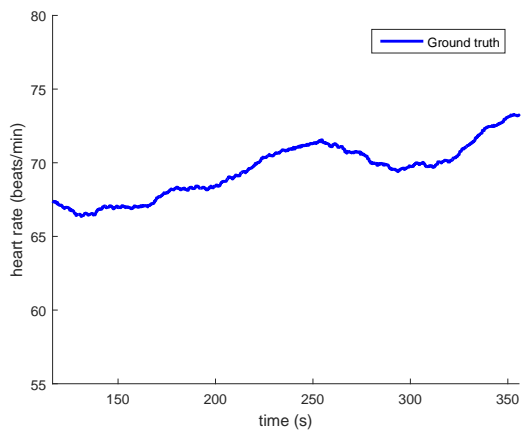}
		\caption{ Heart rate ground truth for exercise data using 1 minute sliding window from finger sensor signal, subject NO. 10. Compared with the resting heart rate shown in Fig. \ref{fig:rate_DLFUMI}, the heart rate after exercise is much higher and maintains larger variability. }\label{fig:HB_rate_GT_exercise}
	\end{center}
\end{figure}

\subsection{Heartbeat Detection and Rate Estimation on Exercise Data}\label{sec:exp_exercise}
{In this section, the heart rate estimation was performed on exercise data collected after each subject spent time bicycling. The goal of this experimental setting is to show the robustness of the proposed algorithm on BCG signals with perturbation and high heart rate variabilities. Before the data collection, each subject was asked to pedal a stationary upright bicycle for two minutes. After the exercise, the subject was asked to lie flat on the back until the subject's blood pressure went back to normal. Then the BCG signals and ground truth information was collected again as described in Sec. \ref{sec:3_sensor} and \ref{sec:data_desc}.  Fig. \ref{fig:BCG_sig_exercise} shows the exemplary BCG signal segment after exercise for subject No. 10, where we can see compared with the resting data shown in Fig. \ref{fig:BCG_FS_plot}, exercise data maintains much larger signal magnitude and shorter J-J perk interval. Fig. \ref{fig:HB_rate_GT_exercise} plots the heart rate ground truth for subject No. 10 after exercise using 1 minute sliding window from finger sensor signal, where it clearly shows the heart rate after exercise is much higher and maintains larger variability in comparison with the heart rate of the resting signal shown in Fig. \ref{fig:rate_DLFUMI}. }

\begin{table}
	\caption{Heart rate estimation performance from BCG exercise data, bold for the best, underline for the second best.} \centering
	\begin{tabular}
		{ c c }\hline
		Alg.  & Overall MAE (beat/min.) \\\hline
		WPPD & 1.88   \\
		CA& 1.90  \\
		EN  &4.13\\
		\underline{HT} & \underline{1.13}\\
		\textbf{DL-FUMI}& 	\textbf{1.01} \\\hline
	\end{tabular}
	\label{tab:HB_rate_exercise}
\end{table}

{To show the robustness of estimated heartbeat concept by proposed algorithm that is able to account for large perturbation and variability during sleeping, the heartbeat estimated individually from each subject in Sec. \ref{sec:exp_indivi_training} is directly applied to the exercise data to obtain the confidence value from the hybrid detector. Then a frequency analysis discussed in Sec. \ref{sec:exp_batch_training} is performed for heart rate estimation and compared with the WPPD, CA, EN and HT algorithms. The overall performance (MAE) is presented in Table. \ref{tab:HB_rate_exercise}, where it can be seen DL-FUMI still achieved the best performance on BCG exercise signal with  perturbation and high heart rate variabilities. For some unexpected reason, the exercise data for Subject No. 9 was not collected. So the overall performance shown in the Table \ref{tab:HB_rate_exercise} is the MAE over 39 subjects except subject No. 9. }

\begin{figure}
	\begin{center}
		\includegraphics[width=9cm]{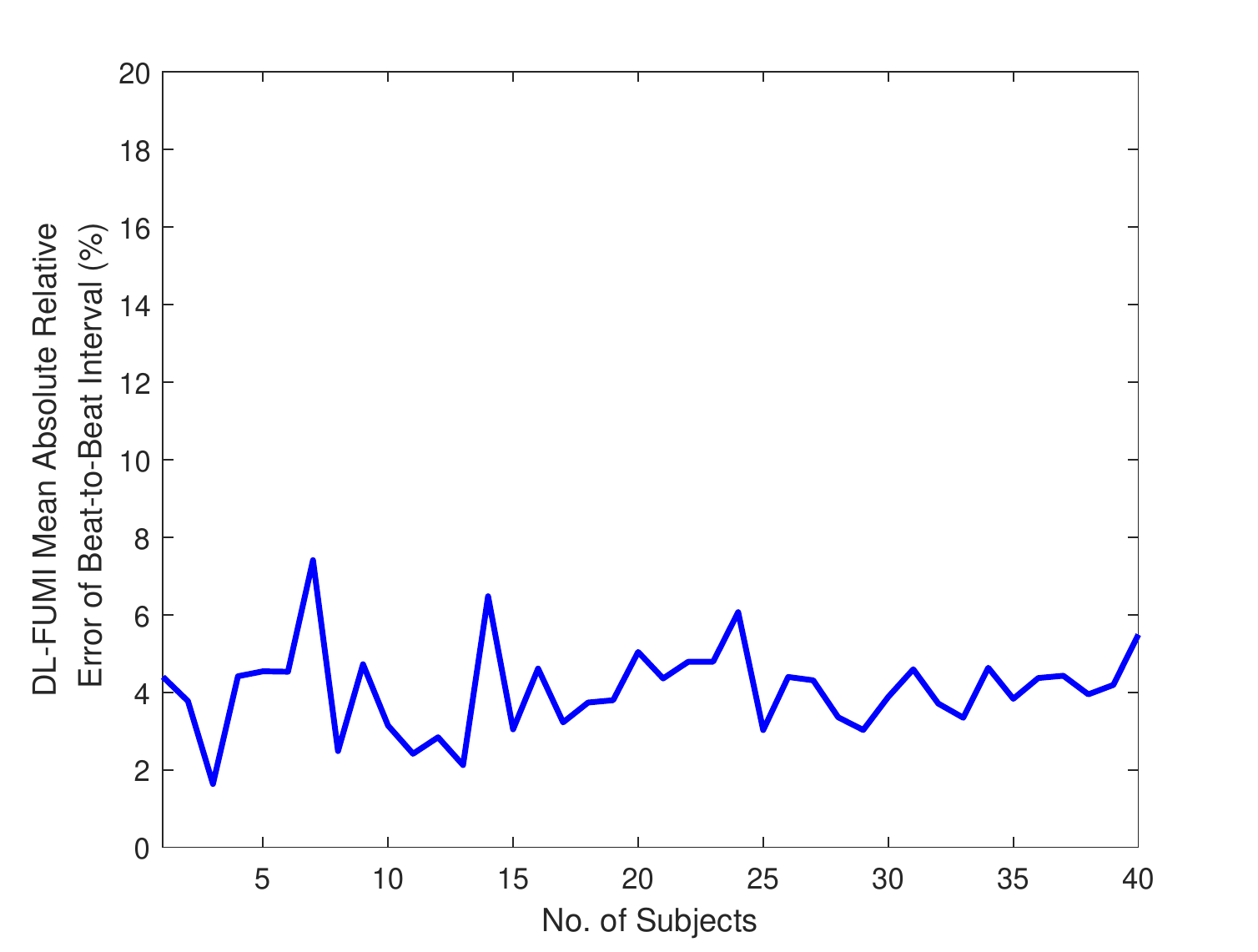}
		\caption{DL-FUMI mean absolute relative error of beat-to-beat interval, the overall relative error for the forty subjects is $4.07\%$.}\label{fig:B2B_error_plot}
	\end{center}
\end{figure}

\begin{figure*}[!hbt]
	\begin{footnotesize}
		\begin{eqnarray}
		\mathbf{d}_k^-&=&\left\{\sum_{i=1}^{N^+} \left[P(z_i=1)\psi\alpha_{ik}(\mathbf{x}_i - \sum_{t=1}^T\alpha_{it}\mathbf{d}_t^+-\sum_{l=1,l\neq k}^M\alpha_{il}\mathbf{d}_l^-)+ P(z_i=0)\psi\alpha_{ik}(\mathbf{x}_i - \sum_{l=1,l\neq k}^M\alpha_{il}\mathbf{d}_l^-)\right]+\sum_{i=1}^{N^-} \left[\alpha_{ik}(\mathbf{x}_i - \sum_{l=1,l\neq k}^M\alpha_{il}\mathbf{d}_l^-)\right]\right. \nonumber \\
		&&\left.-\Gamma\sum_{t=1}^T\cos\theta_{kt}\mathbf{d}_{t^{old}}^+\right\}\left\{\sum_{i=1}^{N+}\psi\alpha_{ik}^2+\sum_{i=1}^{N-}\alpha_{ik}^2\right\}^{-1}
		\label{eqn:update_dk}
		\end{eqnarray}
		\hrulefill
	\end{footnotesize}
\end{figure*}

{A distinct advantage of DL-FUMI over the average heart-rate methods WPPD, CA, EN and HT is its ability to provide beat-to-beat heart rate estimates.  To illustrate, we form the difference of an individual beat-to-beat interval (BBI) obtained from DL-FUMI and GT, take the absolute value, divide by the BBI of GT, and average over the number of beat intervals.  We call this measure as the mean absolute relative error of BBI.  The results averaged from 5 individual runs for the 40 subjects are shown in Fig. \ref{fig:B2B_error_plot}, where DL-FUMI is trained for each subject separately.  There is some variation of the relative error among the 40 subjects, the highest error is less than 8\% and the lowest less than 2\%. The mean of the relative error over the 40 subjects is only $4.07\%$.  }

\section{CONCLUSION }\label{sec:8_conclusion}
In this paper,  heartbeat characterization was formulated as a MIL problem and addressed using the DL-FUMI algorithm. Experimental results show that with enough training data, the algorithm is able to perform well in heartbeat characterization and heart rate estimation with average error less than 1 beat/min. We plan to apply DL-FUMI to the long-term monitored data acquired at TigerPlace, an active aging-in-place retirement community developed by the MU Sinclair School of Nursing and CERT at the University of Missouri \cite{rantz2015new}, and study about how the position, posture, and body movement affect the performance of the proposed algorithm.

%


%

\section{Derivation of DL-FUMI update equations}
\label{sec:gFUMI_update}

This section provides a derivation of DL-FUMI update equations. When updating the dictionary $\mathbf{D}$, the sparse weights $\left\{\boldsymbol{\alpha}_i\right\}_{i=1}^{N}$ are held fixed.
To update one of the atoms in $\mathbf{D}$, \eqref{eqn:E_gFUMI} is minimized with respect to the corresponding atom while keeping all other atoms constant. The resulting update equations for $\mathbf{d}_{t}^+$ and $\mathbf{d}_{k}^-$  are shown in   \eqref{eqn:update_dt} and \eqref{eqn:update_dk}.  

\begin{footnotesize}
	\begin{equation}
	\mathbf{d}_t^+=\frac{\sum_{i=1}^{N^+} \left[P(z_i=1)\alpha_{it}(\mathbf{x}_i - \sum_{l=1,l\neq t}^T\alpha_{il}\mathbf{d}_l^+-\sum_{k=1}^M\alpha_{ik}\mathbf{d}_k^-)\right]}{\sum_{i=1}^{N+}\left[P(z_i=1)\alpha_{it}^2\right]}
	\label{eqn:update_dt}
	\end{equation}
\end{footnotesize}

Note,  $P(z_i | \mathbf{x}_i, \boldsymbol{\theta}^{(t-1))}$ is denoted as $P(z_i)$ for simplicity.

When updating the sparse weights,  $\left\{\boldsymbol{\alpha}_i\right\}_{i=1}^N$, it should be noted that the sparse weight vector $\boldsymbol{\alpha}_i$ for instance $\mathbf{x}_i$ is not dependent on any other instances.
%
%

The gradient with respect to $\boldsymbol{\alpha}_i$ without considering the $l_1$ penalty term is:
\begin{small}
	\begin{eqnarray}
	&&\frac{\partial F^+}{\partial \boldsymbol{\alpha}_i}=-\begin{bmatrix}P(z_i=1)\mathbf{D}^+ & \mathbf{D}^-\end{bmatrix}^T\mathbf{x}_i+\left( P(z_i=1)\mathbf{D}^T\mathbf{D}\right.\nonumber \\
	&&\left.+P(z_i=0)\begin{bmatrix}\mathbf{0}_{d \times T} & \mathbf{D}^-\end{bmatrix}^T\begin{bmatrix}\mathbf{0}_{d \times T} & \mathbf{D}^-\end{bmatrix} \right)\boldsymbol{\alpha}_i.
	\label{eqn:gradient_alpha_plus}
	\end{eqnarray}
\end{small}
Then $\boldsymbol{\alpha}_i$ at $l^{th}$ iteration can be updated using gradient descent,

\begin{equation}
\boldsymbol{\alpha}_i^{l}=\boldsymbol{\alpha}_i^{l-1}-\eta_i\frac{\partial F^+}{\partial \boldsymbol{\alpha}_i},
\label{eqn:alpha_plus_update}
\end{equation}
followed by a soft-thresholding:
\begin{eqnarray}
\left\{ \begin{array}{l}
\boldsymbol{\alpha}_{i}^{l+}=S_{\lambda P(z_i=1)}\left(\boldsymbol{\alpha}_i^{l+}\right) \\
\boldsymbol{\alpha}_{i}^{l-}=S_{\lambda}\left(\boldsymbol{\alpha}_i^{l-}\right)
\label{eqn:sf_alpha_pos}
\end{array}\right.,
\end{eqnarray}
s.t. $S_{\lambda}\left(\mathbf{x}[i]\right)=sign(\mathbf{x}[i])\max(|\mathbf{x}[i]|-\lambda,0)$, $i=1,...,d$.

Following a similar proof to that in \cite{facchinei2007finite}, when \begin{scriptsize}$\eta_i\in\left(0, \left(\lambda_{max}\left(P(z_i=0)\left[\mathbf{0}_{d \times T} \;  \mathbf{D}^-\right]^T\left[\mathbf{0}_{d \times T} \; \mathbf{D}^-\right]+P(z_i=1)\mathbf{D}^T\mathbf{D}\right)\right)^{-1}\right)$\end{scriptsize}, the update of $\boldsymbol{\alpha}_i$ using a gradient descent method with step length $\eta_i$ monotonically decreases the value of the objective function, where $\lambda_{max}(\mathbf{A})$ denotes the maximum eigenvalue of $\mathbf{A}$. For simplicity,  $\eta$ was set as $\eta=\frac{1}{\lambda_{max}\left( \mathbf{D}^T\mathbf{D}\right)}$ for all $\boldsymbol{\alpha}_i$, $\mathbf{x}_i\in\mathbf{B}_j^+$.
%
%
%
%

A similar update can be used for points from negative bags.  The resulting update equation for negative points is:

\begin{small}
	\begin{equation}
	 \boldsymbol{\alpha}_i^l=S_{\lambda}\left(\boldsymbol{\alpha}^{l-1}_i+\frac{1}{\lambda_{max}\left(\mathbf{D}^{-T}\mathbf{D}^-\right)}\left(\mathbf{D}^{-T}(\mathbf{x}_i-\mathbf{D}^-\boldsymbol{\alpha}^{l-1}_i)\right)\right)
	\label{eqn:alpha_minus_update}
	\end{equation}
\end{small}

The sparse weights corresponding to target dictionary atoms are set to 0 for all points in negative bags.

{  \bibliographystyle{IEEEtran}
	\bibliography{DLFUMI_HB_journal_Ref}}

\end{document}